\documentclass[conference]{IEEEtran}
\usepackage{fancyhdr}
\usepackage{cite}
\usepackage{amsmath,amssymb,amsfonts}
\usepackage{graphicx}
\usepackage{textcomp}
\usepackage{xcolor}
\usepackage{algorithm}
\usepackage[noend]{algorithmic}
\usepackage{booktabs}
\usepackage{svg}
\usepackage{url}
\usepackage{xspace}
\usepackage{subcaption}
\usepackage{enumitem}
\usepackage{siunitx}
\usepackage{multirow}
\usepackage{tikz}
\usetikzlibrary{
  arrows.meta,
  calc,
  fit,
  backgrounds,
  positioning,
  shapes.geometric,   
  shadows.blur        
}
\usepackage{amsmath, amssymb}

\definecolor{initbg}{RGB}{187,222,251}    
\definecolor{durbg}{RGB}{255,224,178}     
\definecolor{embg}{RGB}{200,230,201}      
\definecolor{transbg}{RGB}{225,190,231}   
\definecolor{deltabg}{RGB}{255,205,210}   
\definecolor{brickbg}{RGB}{178,223,219}   

\newcommand{\tensoc}{\textsc{Tens-1c}\xspace}
\newcommand{\tensmc}{\textsc{Tens-mc}\xspace}
\newcommand{\tensgpu}{\textsc{Tens-gpu}\xspace}
\newcommand{\tenspy}{\textsc{Tens-Py}\xspace}
\newcommand{\baseoc}{\textsc{Base-1c}\xspace}
\newcommand{\basemc}{\textsc{Base-mc}\xspace}

\newcommand{\hlmath}[2]{\colorbox{#1}{$\displaystyle #2$}}

\def\BibTeX{{\rm B\kern-.05em{\sc i\kern-.025em b}\kern-.08em
    T\kern-.1667em\lower.7ex\hbox{E}\kern-.125emX}}

\newcommand{\algcomment}[1]{\hfill{\color{gray}\footnotesize\textit{// #1}}}
\newcommand{\algcommentlong}[1]{{\color{gray}{\footnotesize\textit{// #1}}}}
\newcommand{\phase}[2]{%
  \item[] \makebox[\linewidth][c]{%
    \rule[0.5ex]{3em}{0.4pt}\quad\textsc{#1}\hspace{1em}#2\quad\rule[0.5ex]{3em}{0.4pt}%
  }%
}
\newcommand{\phasesimple}[1]{%
  \item[] \makebox[\linewidth][c]{%
    \rule[0.5ex]{3em}{0.4pt}\quad\textsc{#1}\quad\rule[0.5ex]{3em}{0.4pt}%
  }%
}

\begin{document}

\title{High-Performance Tensor Formulation of the Viterbi Algorithm for Hidden Semi-Markov Models}

\author{\IEEEauthorblockN{Lorenzo Piarulli}
\IEEEauthorblockA{\textit{Department of Computer Science} \\
\textit{Sapienza University of Rome}\\
piarulli@di.uniroma1.it}
\and
\IEEEauthorblockN{Elia Belli}
\IEEEauthorblockA{\textit{Department of Computer Science} \\
\textit{Sapienza University of Rome}\\
belli.2006305@studenti.uniroma1.it}
\and
\IEEEauthorblockN{Daniele De Sensi}
\IEEEauthorblockA{\textit{Department of Computer Science} \\
\textit{Sapienza University of Rome}\\
desensi@di.uniroma1.it}
\and
}

\maketitle
\thispagestyle{fancy}
\lhead{}
\rhead{}
\chead{}
\lfoot{\footnotesize{
SC26, November 15-20, 2026, Chicago, Illinois, USA
\newline 979-8-3195-4789-7/26/\$31.00 \copyright 2026 IEEE}}
\rfoot{}
\cfoot{}
\renewcommand{\headrulewidth}{0pt}
\renewcommand{\footrulewidth}{0pt}

\begin{abstract}
Hidden Semi-Markov Models (HSMMs) are fundamental probabilistic models widely adopted across diverse domains, from computational biology to finance and signal processing. The Viterbi algorithm decodes the most likely state sequence given an HSMM and can be applied iteratively for ab initio model learning. However, existing Viterbi implementations remain sequential, and GPU-accelerated solutions are entirely absent, making HSMM decoding impractical for large-scale workloads. We present a tensor-based formulation of the Viterbi algorithm for HSMMs, restructuring the inner loops into tensor operations that naturally map onto SIMD units and massively parallel architectures. Building on this formulation, we provide optimized implementations spanning single- and multi-core CPUs, and, for the first time, GPU. Experimental evaluation demonstrates speedups of up to 14$\times$ on a single core, over 200$\times$ with multi-core, and over 570$\times$ on GPU over the state-of-the-art sequential baseline, establishing a new performance baseline for large-scale HSMM decoding.
\end{abstract}


\section{Introduction}
\label{sec:introduction}

High-Performance Computing (HPC) architectures are evolving at an extraordinary rate. Over the past decades, we have transitioned from CPU-only computation to massive multicore processors, and then to GPUs. Now, driven by the rise of artificial intelligence, entirely new accelerator architectures are emerging, including dataflow engines, systolic arrays, and SIMD-centric designs, that promise unprecedented throughput for structured, regular computations. Yet, while hardware evolves rapidly, the algorithms that run on it do not always keep up. Application scientists tend to be conservative: their software frameworks are enormously complex, and restructuring a working codebase to exploit a new architecture is a daunting, error-prone endeavor. In many cases, the cost and complexity of migration simply outweigh the perceived benefit, and teams understandably choose to keep a working pipeline rather than risk breaking it for uncertain gains. As a result, many fundamental algorithms, including those that today underpin astonishing scientific discoveries, remain anchored to decades-old sequential formulations that hide parallelization possibilities and, consequently, remain confined to single-threaded CPU execution. Worse still, practitioners often resort to simplified or truncated versions of their models simply because the full, general formulation would be computationally infeasible on the sequential hardware.



\iftrue
Hidden Semi-Markov Models (HSMMs) are versatile probabilistic frameworks with applications spanning diverse fields, from genome annotation and segmentation in computational biology~\cite{gabriel2024braker3} to finance~\cite{qin2024hsmmfinance}, speech recognition~\cite{zenspeech}, and signal processing~\cite{YU2010215}. As a generalization of the classical \textbf{Hidden Markov Model (HMM)}, an HSMM describes a system transitioning through a finite set of hidden states at discrete time intervals. In this paradigm, the system's internal state remains unobservable, manifesting only through visible emissions governed by state-specific probabilities, while state progression is regulated by a fixed transition matrix.

A practical illustration of this framework is found in sleep-cycle prediction from physiological data: heartbeat measurements serve as the observations, while the underlying sleep stages represent the hidden states to be inferred. Similarly, in HPC applications like genome annotation, a DNA sequence is modeled as a series of nucleotides (observations); the objective is to classify each nucleotide as belonging to either a coding or non-coding region (hidden states). Crucially, these systems often exhibit temporal persistence, remaining in a specific state for numerous consecutive time steps before transitioning, a characteristic that motivates the use of state-duration modeling.

Standard HMMs, however, are fundamentally limited in capturing these temporal dynamics, as state occupancy is inherently restricted to a geometric, memoryless distribution. \textbf{HSMMs} generalize this framework by permitting each hidden state to persist for a variable interval, explicitly defined by a duration distribution~\cite{YU2010215}. Consequently, the model incorporates three core probabilistic elements: state transitions, which govern the likelihood of moving between states; state-specific durations, which model the time spent within a state; and observation emissions, which define the probability of an observation given the current state. 


HSMMs are particularly indispensable in computational biology for tasks such as genome annotation and segmentation~\cite{gabriel2024braker3,korf2004}. Once a genome has been sequenced, a central challenge lies in identifying which nucleotide sequences correspond to protein-coding exons, non-coding introns, regulatory elements, or intergenic regions. Interpreting the nucleotide sequence as an observation sequence, genome annotation amounts to inferring the hidden functional category of each region. The key difficulty is that genomic features can span from a few dozen to many thousands of nucleotides; modeling such extended segments requires explicit duration distributions that bypass the constant transition probability inherent to HMMs, making HSMMs a natural fit. This utility extends to chromatin state annotation, \texttt{CpG} island detection, and other problems where segment lengths carry vital biological meaning~\cite{ernst2012chromhmm, durbin1998biological}.


Given a sequence of observations, two central tasks arise: \emph{learning}, which estimates the model parameters from observed data, and \emph{decoding}, which identifies the most likely hidden state sequence. Three fundamental algorithms solve these tasks: the \emph{Forward--Backward} algorithm~\cite{rabiner, YU2010215} computes state probabilities at each time step, the \emph{Baum--Welch} algorithm~\cite{baumwelch, rabiner, YU2010215}, an instance of Expectation--Maximization~\cite{dempster1977em}, estimates model parameters, and the \emph{Viterbi} algorithm performs decoding. The Viterbi algorithm directly solves the genome annotation problem; moreover, by iteratively applying Viterbi decoding and re-estimating parameters from the decoded sequences, one can implement a Viterbi training loop, a technique widely used for \emph{ab initio} gene prediction~\cite{lomsadze2005} when no prior data is available. Since the Viterbi algorithm serves both decoding and \emph{ab initio} learning, it is the most widely adopted of the three, making it a primary candidate for acceleration.

However, while the Viterbi algorithm for a standard HMM has a complexity of $O(TN^2)$, the HSMM formulation introduces an additional loop over all possible durations, raising the complexity to $O(TN^2D)$, where $D$ is the maximum admissible duration. In practical applications where $D$ ranges from hundreds to thousands~\cite{korf2004}, this extra dimension poses a major computational bottleneck. Despite the critical importance of these problems, the algorithmic formulations used for HSMM inference have remained largely unchanged since their original proposal. 


Historically, the classical Viterbi algorithm for HSMMs has been implemented via four nested loops~\cite{YU2010215, rabiner, gedon01092003}. Despite the proliferation of HPC resources, existing implementations remain confined to scalar, single-threaded CPU execution~\cite{berard2025review, du2014biomvrhsmm, hsmmr, hsmmlearn}, in stark contrast to standard HMMs, which benefit from an extensive ecosystem of GPU-accelerated and SIMD-optimized frameworks~\cite{firtina2024aphmm, hmmer3, yu2014gpu, Jiang2018, Hassan2021}.


Consequently, researchers face a restrictive trade-off: either endure prohibitively long execution times or artificially truncate their datasets, preventing the full expressive potential of HSMMs from being realized in large-scale scientific workflows.

Given the importance of HSMMs and the widening gap between the computational demands of real-world applications and the capabilities of existing sequential implementations, this work proposes the following contributions:

\begin{enumerate}[leftmargin=*]
    \item We introduce a novel \textbf{Tensor-Based formulation of the Viterbi algorithm for Hidden Semi-Markov Models}. By restructuring the traditional three-nested inner loops into dense tensor operations, our approach naturally maps onto the SIMD and SPMD execution models of modern HPC architectures. This reformulation exposes significant optimization opportunities that were previously inaccessible in sequential scalar implementations.

    \item We leverage this formulation to deliver optimized implementations across CPUs utilizing SIMD vectorization and threading, and, for the first time, GPUs. To facilitate broader adoption, all implementations are released as a high-performance open-source library designed for seamless integration into existing scientific workflows.

    \item We conduct an extensive performance evaluation across three CPU and five GPU architectures, demonstrating speedups of up to $570\times$ over state-of-the-art HSMM frameworks. 
\end{enumerate}

\else
Hidden Semi-Markov Models (HSMMs) are versatile probabilistic frameworks with applications spanning diverse fields, from genome annotation and segmentation in computational biology~\cite{gabriel2024braker3} to finance~\cite{qin2024hsmmfinance}, speech recognition~\cite{zenspeech}, and signal processing~\cite{YU2010215}. As a generalization of the classical \textbf{Hidden Markov Model (HMM)}, an HSMM builds upon a core architecture where a system transitions through a finite set of hidden states at discrete time intervals. In this paradigm, the system's internal state remains unobservable, manifesting only through visible emissions governed by state-specific probabilities, while state progression is regulated by a fixed transition matrix.

A practical illustration of this framework is found in sleep-cycle prediction from physiological data: heartbeat measurements serve as the observations, while the underlying sleep stages represent the hidden states to be inferred. Similarly, in high-performance computing (HPC) applications like genome annotation, a DNA sequence is modeled as a series of nucleotides (observations); the objective is to classify each nucleotide as belonging to either a coding or non-coding region (hidden states). Crucially, these systems often exhibit temporal persistence, remaining in a specific state for numerous consecutive time steps before transitioning, a characteristic that motivates the use of state-duration modeling.

Standard HMMs, however, are fundamentally limited in capturing these temporal dynamics, as state occupancy is inherently restricted to a geometric, memoryless distribution. \textbf{HSMMs} generalize this framework by permitting each hidden state to persist for a variable interval, explicitly defined by a duration distribution~\cite{YU2010215}. Consequently, the model incorporates three core probabilistic elements: state transitions, which govern the likelihood of moving between states; state-specific durations, which model the time spent within a state; and observation emissions, which define the probability of an observation given the current state. Although this flexibility greatly enhances the model's expressive power, it significantly increases the computational burden of inference. While the Viterbi algorithm for a standard HMM scales as $O(TN^2)$—where $T$ is the sequence length and $N$ the number of hidden states—the HSMM formulation introduces an additional loop over all possible durations. This elevates the complexity to $O(T N^2D)$, where $D$ is the maximum admissible duration. In practical applications where $D$ may range from hundreds to thousands of steps, this extra dimension poses a major computational bottleneck.

HSMMs are particularly valuable in computational biology for genome annotation and segmentation~\cite{gabriel2024braker3,korf2004}. Once a genome has been sequenced, a central challenge is to determine the functional role of each region: which regions of nucleotides correspond to protein-coding exons, which are non-coding introns, which are regulatory elements such as promoters or enhancers, and which belong to intergenic or repetitive regions. If we interpret the nucleotide sequence as an observation sequence, where each nucleotide represents a single observation, then annotating the genome amounts to inferring the hidden state, that is, the functional category, associated with each region. The key difficulty is that genomic functional regions, such as exons or introns, can span from a few tens to many thousands of nucleotides, thus modeling such extended segments requires explicit duration distributions, making HSMMs a natural and powerful fit for this task. Beyond gene structure prediction, HSMMs have also been applied to chromatin state annotation, \texttt{CpG} island detection, and other problems where segment lengths carry biological meaning~\cite{ernst2012chromhmm, durbin1998biological}.

Given a sequence of observations, two central tasks arise: \emph{decoding}, which asks what hidden state sequence most likely generated the observations, and \emph{learning}, which asks, given both observations and their associated state sequences, what HSMM best describe the underlying process. Three fundamental algorithms address these questions. The \emph{Forward--Backward} algorithm~\cite{rabiner, YU2010215} computes the probability of each hidden state at each time step, given the full observation sequence. The \emph{Baum--Welch} algorithm~\cite{baumwelch, rabiner, YU2010215}, an instance of Expectation--Maximization~\cite{dempster1977em}, estimates the model parameters, namely transition, emission, and duration distributions, from observed data, constituting the parameter estimation phase. Finally, the \emph{Viterbi} algorithm performs \emph{decoding}: given a trained model and an observation sequence, it finds the single most likely sequence of hidden states. It is the Viterbi algorithm that directly solves the genome annotation problem. Moreover, by iterating Viterbi decoding and re-estimating the model parameters from the decoded state sequences, one can implement a Viterbi training loop, a technique widely used in genomic frameworks for \emph{ab initio} gene prediction~\cite{lomsadze2005} when no prior data is available for the organism under study. Since the Viterbi algorithm serves both decoding and \emph{ab initio} learning, it is the most widely adopted of the three, making it a natural candidate for acceleration.

However, despite the critical importance of these problems, the algorithmic formulations used for HSMM inference have remained largely unchanged since their original proposal. The classical Viterbi for HSMMs is implemented as four nested loops, iterating over time steps, current and previous states, and possible durations~\cite{YU2010215, rabiner, gedon01092003}. Existing implementations are designed for scalar, single-threaded CPUs, and SIMD optimization has not been explored in depth~\cite{berard2025review, du2014biomvrhsmm, hsmmr, hsmmlearn}. This stands in stark contrast with standard HMMs, for which GPU-accelerated and SIMD-optimized implementations have been developed and studied extensively~\cite{firtina2024aphmm, hmmer3, yu2014gpu, Jiang2018, Hassan2021}. The result is that genomic scientists who rely on HSMM-based frameworks must either wait for prohibitively long runtimes or truncate their inputs, reducing sequence lengths, capping maximum durations, or simplifying the state space, to make computation tractable on current hardware.

Given the importance of HSMMs and the widening gap between the computational demands of real-world applications and the capabilities of existing sequential implementations, this work proposes the following contributions:

\begin{enumerate}
    \item We introduce a novel \textbf{Tensor-Based formulation of the Viterbi algorithm for Hidden Semi-Markov Models} that restructures the sequential 3-inner loops into dense tensor operations that naturally map onto the SIMD and SPMD execution models of modern HPC architectures. This reformulation exposes new optimization opportunities that can be leveraged on both CPU and GPU.

    \item We exploit the optimizations exposed by the tensor-based formulation, and provide efficient implementations of this tensor-based Viterbi algorithm on: \textbf{single-core CPU} with SIMD vectorization, \textbf{multi-core CPU} with OpenMP threading and SIMD, and, for the first time, a \textbf{GPU implementation}. We evaluate the algorithm on X different CPUs and Y different GPUs, reaching speedups of up to $K\times$ over existing HSMM frameworks. All implementations are released as an open-source library designed for easy integration into existing scientific workflows.

\end{enumerate}
\fi

\section{Sequential HSMM Viterbi Algorithm}


We now formalize the elements that define a Hidden Semi-Markov Model. An HSMM is represented by the parameter tuple~\cite{YU2010215, rabiner, gedon01092003}:
\begin{equation}\label{eq:hsmm_params}
    \lambda = \bigl(\mathcal{S},\;
                     \mathcal{O},\;
                     \mathbf{A},\;
                     \mathbf{P},\;
                     \mathbf{B},\;
                     \boldsymbol{\pi}\bigr)\,,
\end{equation}
where $\mathcal{S} = \{s_1, \dots, s_N\}$ is a finite set of~$N$ hidden states, $\mathcal{O} = \{o_1, \dots, o_M\}$ is a finite set of~$M$ observation symbols, and $\boldsymbol{\pi}^{N}[j] = \pi_j$ is the initial probability of state $s_j$. The collection of all transition probabilities forms the \emph{transition probability matrix} $\mathbf{A}^{N \times N}$, where each entry $\mathbf{A}[i,\,j] = a_{ij}$ represents the probability of transitioning from state~$s_i$ to state~$s_j$. Similarly, the collection of all duration probabilities forms the \emph{duration probability matrix} $\mathbf{P}^{N \times D}$, where each entry $\mathbf{P}[j,\,d] = p_j(d)$ represents the probability that state~$s_j$ persists for exactly~$d$ consecutive time-steps, with $d \in \{1, \dots, D\}$. Finally, the emission probabilities define the \emph{emission probability matrix} $\mathbf{B}^{N \times M}$, where $\mathbf{B}[j,\,o] = b_j(o)$ is the probability that state~$s_j$ emits observation~$o$. These matrix definitions of the model parameters will be central to the tensor formulation presented in Sec.~\ref{sec:tensor-based}. Table~\ref{tab:hsmm_components} summarizes all components of the model.




\begin{table}[ht]
\centering
\scriptsize
\caption{Components of an HSMM and Viterbi Algorithm}\label{tab:hsmm_components}
\renewcommand{\arraystretch}{1.2}
\begin{tabular}{@{}llp{5cm}@{}}
\toprule
\textbf{Parameter} & \textbf{Matrix} & \textbf{Description} \\
\midrule
$\mathcal{S} = \{s_1, \dots, s_N\}$
    & $\mathcal{S}^N$
    & Set of $N$ hidden states. \\
$\mathcal{O} = \{o_1, \dots, o_M\}$
    & $\mathcal{O}^M$
    & Set of $M$ observation symbols. \\
$\pi_j$
    & $\boldsymbol{\pi}^{N}$
    & Initial state probability for $s_j$.\\
$a_{ij}$
    & $\mathbf{A}^{N \times N}$
    & Transition prob.\ from $s_i$ to $s_j$; self-transitions governed by zero-duration. \\
$p_j(d)$
    & $\mathbf{P}^{N \times D}$
    & Prob.\ that $s_j$ persists for exactly $d$ steps, $d\in [1,D]$. \\
$b_j(o)$
    & $\mathbf{B}^{N \times M}$
    & Emission prob.\ of observing $o$ in state $s_j$. \\
$\delta_t(j)$
    & $\mathbf{\Delta}^{N \times T}$
    & Likelihood of the most probable sequence ending in $s_j$ at time $t$. \\
$\psi_t(j)$
    & $\mathbf{\Psi}^{N \times T}$
    & Coordinates $(s_i,d)$ of the best predecessor for $s_j$ at time $t$. \\
\bottomrule
\end{tabular}
\end{table}

The Viterbi algorithm is one of the fundamental algorithms for HSMMs. It addresses the following problem: given a model~$\lambda$ and a sequence of observations
$o_0, \dots, o_{T-1}$, the goal is to find the most likely hidden state sequence $\mathbf{s}^* = (q^*_0, \dots, q^*_{T-1})$. To achieve this, for every time step~$t$ and each state $s_j \in \mathcal{S}$, the algorithm computes $\delta_t(j)$, which represents the likelihood of the most probable state sequence ending in state~$s_j$ at time~$t$. Both the sequential and the tensor formulations (Sec.~\ref{sec:tensor-based}) maintain these values in a ${\Delta}^{N \times T}$ matrix:

\begin{equation*}\label{eq:delta_matrix}
    \small\boldsymbol{\Delta}^{N \times T}
    \;=\;
    \begin{bmatrix}
        \delta_1(0)   & \delta_1(1)   & \cdots & \delta_1(T\!-\!1) \\[2pt]
        \delta_2(0)   & \delta_2(1)   & \cdots & \delta_2(T\!-\!1) \\[2pt]
        \vdots        & \vdots        & \ddots & \vdots        \\[2pt]
        \delta_N(0)   & \delta_N(1)   & \cdots & \delta_N(T\!-\!1)
    \end{bmatrix}.
\end{equation*}

Following a dynamic programming approach, the algorithm proceeds in three distinct stages: \textbf{initialization}, \textbf{induction}, and \textbf{backtracking}.

\subsection{Initialization Phase}
\label{sec:seq_init}

The initialization phase covers the first $D$ time steps, where $D$ is the maximum admissible state duration. During this interval, we must account for the possibility that the system has occupied state~$s_j$ since $t=1$ with no prior transition. For each state~$s_j$ and each $1 \leq t \leq D$, the initialization value combines $\hlmath{initbg}{\text{(a)}}$ the initial state probability~$\pi_j$, $\hlmath{durbg}{\text{(b)}}$ the probability that state~$s_j$ persists for exactly~$t$ time steps, and $\hlmath{embg}{\text{(c)}}$ the joint emission probability of observations $o_0, \dots, o_{t-1}$ under state~$s_j$. For $t > D$, no state can have persisted since the beginning, so this contribution is no longer considered. 

\begin{equation}\label{eq:init}
\delta_t(j) = 
\underbrace{\hlmath{initbg}{\pi_j}}_{\text{(a)}}
\;\cdot\;
\underbrace{\hlmath{durbg}{p_j(t)}}_{\text{(b)}}
\;\cdot\;
\underbrace{\hlmath{embg}{\prod_{\tau=0}^{t-1} b_j(o_{t-\tau})}}_{\text{(c)}}
\;,\quad 1 \le t \le D
\end{equation}

\subsection{Induction phase}
\label{sec:seq_inductive}
After initializing the first $D$ time steps, we proceed with the most computationally intensive phase: the induction. For each time step~$t$ and each current state~$s_j$, the goal is to find the previous state~$s_i$ and the duration~$d$ that together maximize the likelihood of reaching~$s_j$ at time~$t$. The formulation is given by Equation~\eqref{eq:induction_delta}, and the corresponding four-nested-loop pseudocode is shown in Algorithm~\ref{alg:sequential-algo}.

The computation combines four factors. Term~\colorbox{deltabg}{(a)} is the previously computed value $\delta_{t-d}(i)$: it encodes the likelihood of the best path ending in state~$s_i$ at time $t\!-\!d$, under the assumption that a transition to~$s_j$ occurred there.  Term~\colorbox{transbg}{(b)} is the transition probability~$a_{ij}$ from state~$s_i$ to state~$s_j$. 

Together, \colorbox{deltabg}{(a)}~and~\colorbox{transbg}{(b)} form the inner maximization: for a fixed duration~$d$, we evaluate all possible source states~$s_i$ and select the one that yields the highest likelihood.

The result is then multiplied by term~\colorbox{durbg}{(c)}, the probability $p_j(d)$ that state~$s_j$ persists for exactly~$d$ consecutive time steps, and by term~\colorbox{embg}{(d)}, the cumulative emission probability of all observations from $t-d+1$ to~$t$ under state~$s_j$. The maximization repeats this for all durations~$d \in \{1, \dots, \min(t,D)\}$ and all source states $s_i$ and selects the best combination. The resulting optimal combination $(d^*,\, i^*)$ for each state~$s_j$ at time step~$t$ is stored in~$\psi_t(j)$, while the corresponding likelihood is stored in~$\delta_t(j)$.

\begin{equation}\label{eq:induction_delta}
    \delta_t(j) =
    \max_{\forall i, \forall d}
    \Bigl[
            \underbrace{
                \hlmath{deltabg}{\delta_{t-d}(i)}
            }_{(a)}
            \cdot
            \underbrace{
                \hlmath{transbg}{a_{ij}}
            }_{(b)}
        \cdot
        \underbrace{
            \hlmath{durbg}{p_j(d)}
        }_{(c)}
        \cdot
        \underbrace{
            \hlmath{embg}{\prod_{k=0}^{d-1} b_j(o_{t-k})}
        }_{(d)}
    \Bigr],
\end{equation}
\begin{algorithm}[ht]
\caption{Sequential HSMM Viterbi Induction.}
\label{alg:sequential-algo}
\footnotesize
\begin{algorithmic}[1]
\smallskip
\FOR{$t = 2$ \TO $T$}
    \FOR{$j = 1$ \TO $N$}
        \STATE $\delta_t(j) \gets -\infty$
        \FOR{$d = 1$ \TO $\min(t,\, D)$}
            \FOR{$i = 1$ \TO $N$}
                \STATE val $\gets$\colorbox{deltabg}{$\delta_{t-d}(i)$}$\cdot$\colorbox{transbg}{$a_{ij}$}$\cdot$\colorbox{durbg}{$p_j(d)$}$\cdot$\colorbox{embg}{$\prod_{k=0}^{d-1} b_j(o_{t-k})$}
                \IF{val $> \delta_t(j)$}
                    \STATE $\delta_t(j) \gets$ val
                    \STATE $\psi_t(j) \gets (d,\, i)$
                \ENDIF
            \ENDFOR
        \ENDFOR
    \ENDFOR
\ENDFOR
\end{algorithmic}
\end{algorithm}

Note that induction starts from $t = 2$; for $t \le D$, the value $\delta_t(j)$ is determined by the maximum of two cases: the system has remained in state $j$ since $t=1$ (initialization), or it transitioned from a previous state $i$ at some time $\tau < t$ (induction). The algorithm takes the greater of these two probabilities.
\subsection{Backtracking Phase}\label{sec:backtracking}

Once $\delta_t(j)$ has been computed $\forall\, t,\; \forall\, j$,
we can recover the optimal state sequence. Starting from the last
time step, we select the state with the highest delta value.
The backtracking then proceeds backwards, until $t = 0$, returning the most likely chain of states. Since we operate in a Semi-Markov
regime, the recovered path will typically exhibit states persisting
across multiple consecutive time steps, reflecting the explicit
duration modeling that distinguishes the HSMM from an HMM.

\section{Tensor-Based Viterbi Algorithm}\label{sec:tensor-based}

The tensor formulation of the Viterbi algorithm follows the same subdivision as the standard one: an \emph{initialization phase}, an \emph{induction phase}, and a \emph{backtracking phase}.
However in this work, we express both initialization and induction using a new tensor formulation.
The backtracking phase remains sequential as it does not represent a computational bottleneck. Instead, optimization efforts focus on the induction phase, where tensor reformulation yields the greatest benefit.

\subsection*{From Loops to Tensors}\label{sec:loop_to_tensor}

As described in the sequential Algorithm~\ref{alg:sequential-algo}, for every time step~$t$ we need to evaluate all possible combinations of a previous state~$s_i$ and a duration~$d$ for each current state~$s_j$.
Formally, for every state $s_j\in\mathcal{S}$ we must consider every pair $(s_i,\,d)$ with $s_i\in\mathcal{S}\setminus\{s_j\}$ and $d\in\{1,\dots,D\}$. Once all possible combinations of $(s_i,d)$ have been computed for each state $s_j \in \mathcal{S}$, our aim is to compute $\delta_t(j)$ for every time step~$t \leq T$.

To compute these combinations and then $\delta_t(j), \forall j$ sequentially, one must iterate over the current states, then over the possible durations, and then again over the previous states, yielding \textbf{three nested loops} already contained inside the main time-step loop. The sequential algorithm therefore requires four nested loops.
\subsection{The \textit{Brick} Representation}\label{sec:brick}

\begin{figure}[h]
    \centering
    \includegraphics[width=0.7\columnwidth]{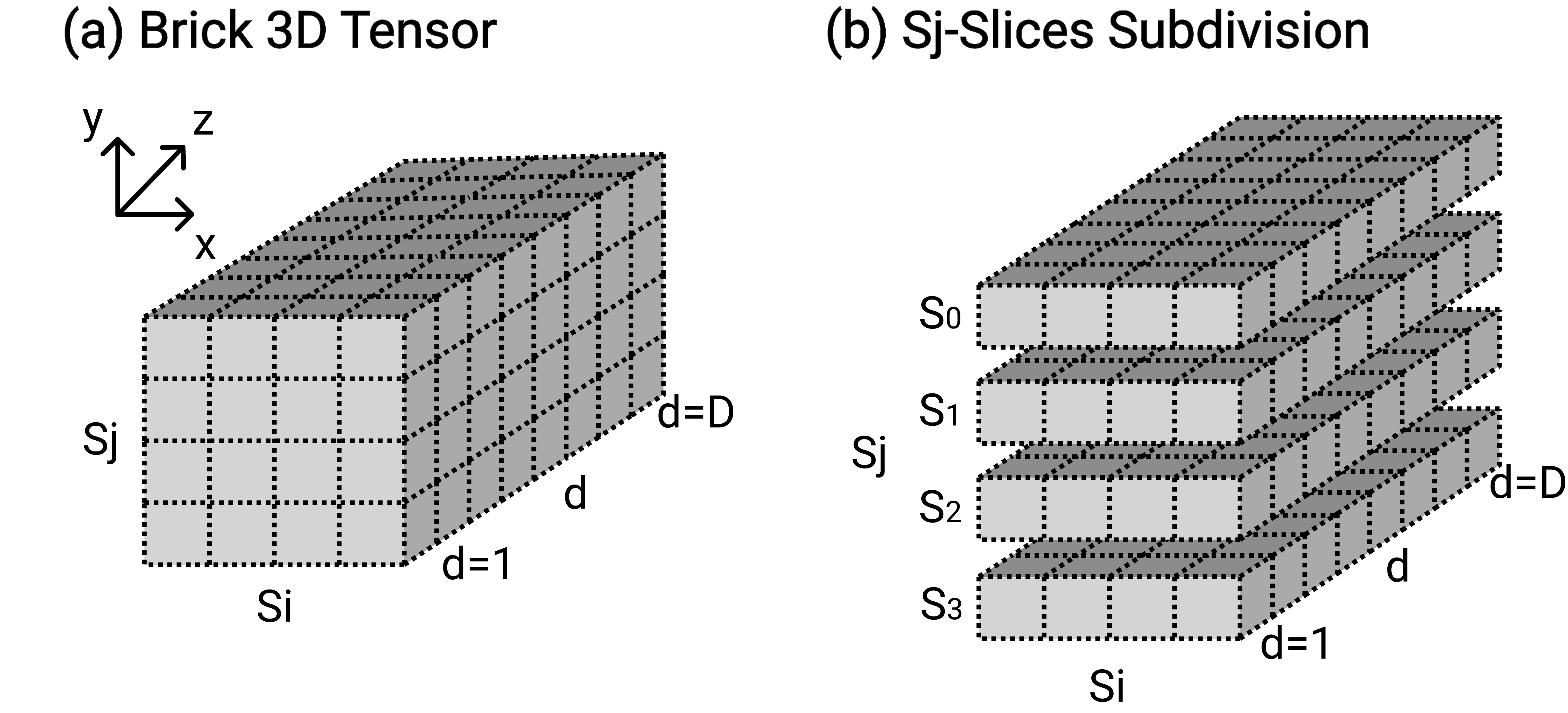}
    \caption{Visualization of the \textit{Brick} 3D tensor and subdivision into $sj$-slices.}
    \label{fig:brick_representation}
\end{figure}

Our key idea is to replace the three nested loops with structured tensor operations that make data reuse explicit and expose independent computations along each axis.
The loops over $(s_j, s_i, d)$ can be naturally mapped onto a three-dimensional tensor, as shown in Figure~\ref{fig:brick_representation}a.
We choose the following layout:

\begin{itemize}
    \item $y$-axis $\;\longrightarrow\;$ target states $s_j$, with $j \in \{1,\dots,N\}$;
    \item $x$-axis $\;\longrightarrow\;$ source states $s_i$, with $i \in \{1,\dots,N\}$;
    \item $z$-axis $\;\longrightarrow\;$ durations $d$, with $d \in \{1,\dots,D\}$.
\end{itemize}

\noindent
With this convention the set of all combinations $(s_j\,,s_i\,,d)$ can be visualized as a 3-dimensional tensor of size $N\times N\times D$ that we will call \textit{Brick} ($\mathcal{B}$), shown in Figure~\ref{fig:brick_representation}a.
Each \emph{slice} of the \textit{Brick} along the $y$-axis corresponds to the complete set of $(s_i,\,d)$ combinations for a single target state~$s_j$, which we call $s_j$-slices (Figure~\ref{fig:brick_representation}b).
Using this representation, we reformulate the sequential induction of Algorithm~\ref{alg:sequential-algo} into three key stages: (i) the \textbf{Brick Construction}, which occurs once (Sec.~\ref{sec:populating}); (ii) the \textbf{Brick Update}, where the \textit{Brick} is modified at each time step $t$ to incorporate temporal factors (as described in Sec.~\ref{sec:time_dep}); and (iii) the \textbf{Maximum Extraction}, which identifies the maximum value and the corresponding coordinates $(s_i, d)$ within each $s_j$-slice. While phases (ii) and (iii) are executed iteratively at each time step, phase (i) is a pre-computation step performed outside the time steps loop. This pipeline, including initialization, takes shape within the Algorithm~\ref{alg:tensor} that will be described line-by-line below.

\begin{algorithm}[ht]
\caption{Tensor-based HSMM Viterbi.}\label{alg:tensor}
\footnotesize{
\begin{algorithmic}[1]
\smallskip
\phase{Initialization}{$(1 \le t \le D)$}
\STATE $\hlmath{embg}{\mathbf{E}^{N\times D}} \gets \textit{Emission Product Computation}$
\STATE $\boldsymbol{\Delta}^{N\times D} \gets \hlmath{initbg}{\boldsymbol{\pi}^{N\times \uparrow}} \odot \hlmath{durbg}{\mathbf{P}^{N\times D}} \odot \hlmath{embg}{\mathbf{E}^{N\times D}}$
\smallskip
\phase{Induction}{$(2 \le t \le T)$}
\STATE $\mathcal{B}_{first}^{N\times N\times D} \gets \hlmath{transbg}{\mathbf{A}^{N\times N \times \uparrow}} \odot \hlmath{durbg}{\mathbf{P}^{N\times \uparrow\times D}}$
\algcomment{Brick Construction}
\smallskip
\FOR{$t = 2$ \TO $T$}
    \smallskip
    \STATE $\hlmath{embg}{\mathbf{E}^{N\times D}} \gets \textit{Emission Product Computation}$
    \smallskip
    \STATE $\hlmath{deltabg}{\boldsymbol{\Delta}_{past}^{N\times D}} 
            \gets \boldsymbol{\Delta}(t - D:t - 1)^{N\times D}$ \algcomment{Past Delta Extraction}
    \smallskip
    \STATE $\hlmath{brickbg}{\mathcal{B}_{(t)}^{N\times N\times D}} \gets \mathcal{B}_{first} \odot \hlmath{deltabg}{\boldsymbol{\Delta}_{past}^{\uparrow\times N\times D}}
            \odot \hlmath{embg}{\mathbf{E}^{N\times \uparrow\times D}}$ \algcommentlong{Brick Update}
    \smallskip
    \FOR{$j = 1$ \TO $N$}
    \STATE $\Psi_j(t)
    \gets
    \arg\max_{d,\,i}\;
    \hlmath{brickbg}{\mathcal{B}_{(t)}[s_j\text{-slice}]^{N\times D}}$
     \algcomment{Max. Extraction}
    \ENDFOR 
\ENDFOR
\smallskip
\phasesimple{Backtracking}
\STATE $\mathbf{q}^* = \textit{Backtracking}(\Psi^{N\times T},\; \Delta^{N\times T})$
\smallskip
\end{algorithmic}
}
\end{algorithm}

\subsection{Brick Construction}\label{sec:populating}

\begin{figure}[h]
    \centering
    \includegraphics[width=0.8\columnwidth]{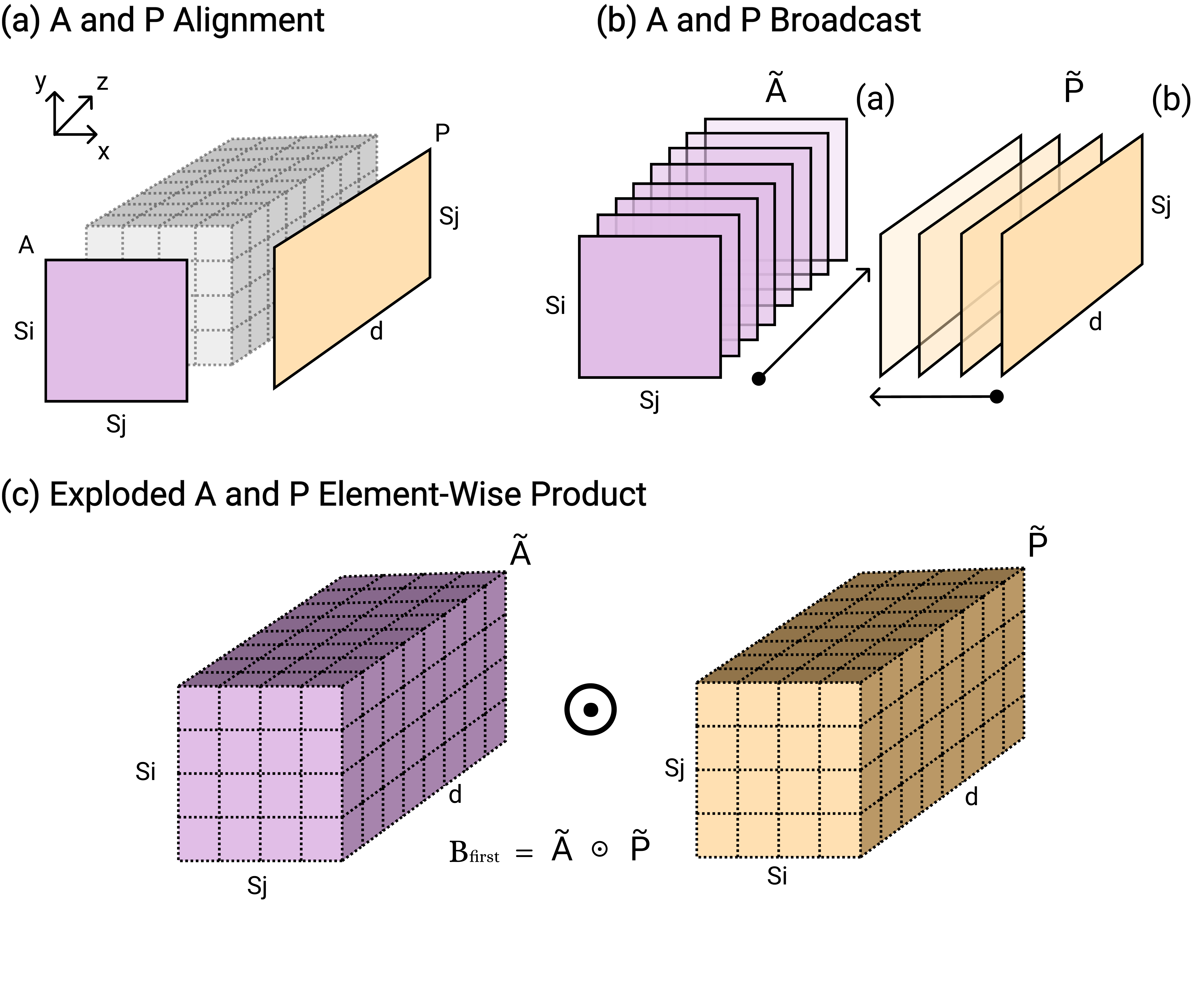}
    \caption{Alignment of $\mathbf{A}$ and $\mathbf{P}$ within the  \textit{Brick} 3D tensor and the corresponding representation of the broadcasted product.}
    \label{fig:broadcasted_product}
\end{figure}

The key observation is the following: the \textit{transition probability matrix} $\mathbf{A}^{N\times N}$ and the \textit{duration probability matrix} $\mathbf{P}^{N\times D}$ do not depend on the time step~$t$. Their contribution to the \textit{Brick} can therefore be precomputed once, before entering the time-step loop. This corresponds to line~3 of Algorithm~\ref{alg:tensor}.
As shown in Figure \ref{fig:broadcasted_product}a, the transition matrix~$\mathbf{A}$ is a two-dimensional $N\times N$ matrix. Within our \textit{Brick}, it occupies the \emph{front face}, i.e.\ it is aligned with the $y$- and $x$-axes.
The duration probability matrix~$\mathbf{P}$ is a two-dimensional $N\times D$ matrix positioned on the \emph{side face}, i.e.\ aligned with the $y$- and $z$-axes (Figure \ref{fig:broadcasted_product}a).

To combine these two matrices into a single three-dimensional tensor, we perform the following product:
\begin{equation}\label{eq:A_D}
    \mathcal{B}_{first}^{N\times N\times D} =
    \underbrace{
        \hlmath{transbg}{\widetilde{\mathbf{A}}^{N\times N\times \uparrow}}
    }_{(a)}
    \;\odot\;
    \underbrace{
        \hlmath{durbg}{\widetilde{\mathbf{P}}^{N\times \uparrow \times D}}
    }_{(b)}
\end{equation}
where the $\uparrow$ symbol denotes the axis along which broadcasting occurs, expanding the tensor to match the dimensions of the corresponding operand. Concretely, as shown in Figure~\ref{fig:broadcasted_product}b-c, the operation can be understood in two steps:

\colorbox{transbg}{(a)}\textit{Broadcast} $\mathbf{A}$: replicate the $N\times N$ matrix $D$ times along the $z$-axis, obtaining a tensor $\widetilde{\mathbf{A}}^{N\times N\times D}$ (Figure~\ref{fig:broadcasted_product}b-(a)). 

\colorbox{durbg}{(b)}\textit{Broadcast} $\mathbf{P}$: replicate the $N\times D$ matrix $N$ times along the $x$-axis, obtaining a tensor $\widetilde{\mathbf{P}}^{N\times N\times D}$ (Figure~\ref{fig:broadcasted_product}b-(b)). 

\noindent
Then, the \textit{Brick} is given by the element-wise product of the two broadcast matrices: $\mathcal{B}_{first}^{N\times N\times D}=\widetilde{\mathbf{A}}^{N\times N\times D}\odot\widetilde{\mathbf{P}}^{N\times N\times D}$ (Figure~\ref{fig:broadcasted_product}c). We refer to this as a broadcasted product: a fundamental operation of our tensor-based Viterbi algorithm. 

\subsection{Brick Update}\label{sec:time_dep}

After constructing the static \textit{Brick}~$\mathcal{B}_{first}$, we enter the main loop over time-steps.
At each step~$t$, two additional matrices must be incorporated: the \emph{past delta values} $\boldsymbol{\Delta}_{past}$ and the \emph{emission probability product} $\mathbf{E}$, both of which depend strictly on~$t$. This phase corresponds to lines~5--7 of Algorithm~\ref{alg:tensor}.

\subsubsection{\textbf{Extracting Past Delta Matrix}}\label{ssec:past_delta}

\begin{figure}[h]
    \centering
    \includegraphics[width=0.8\columnwidth]{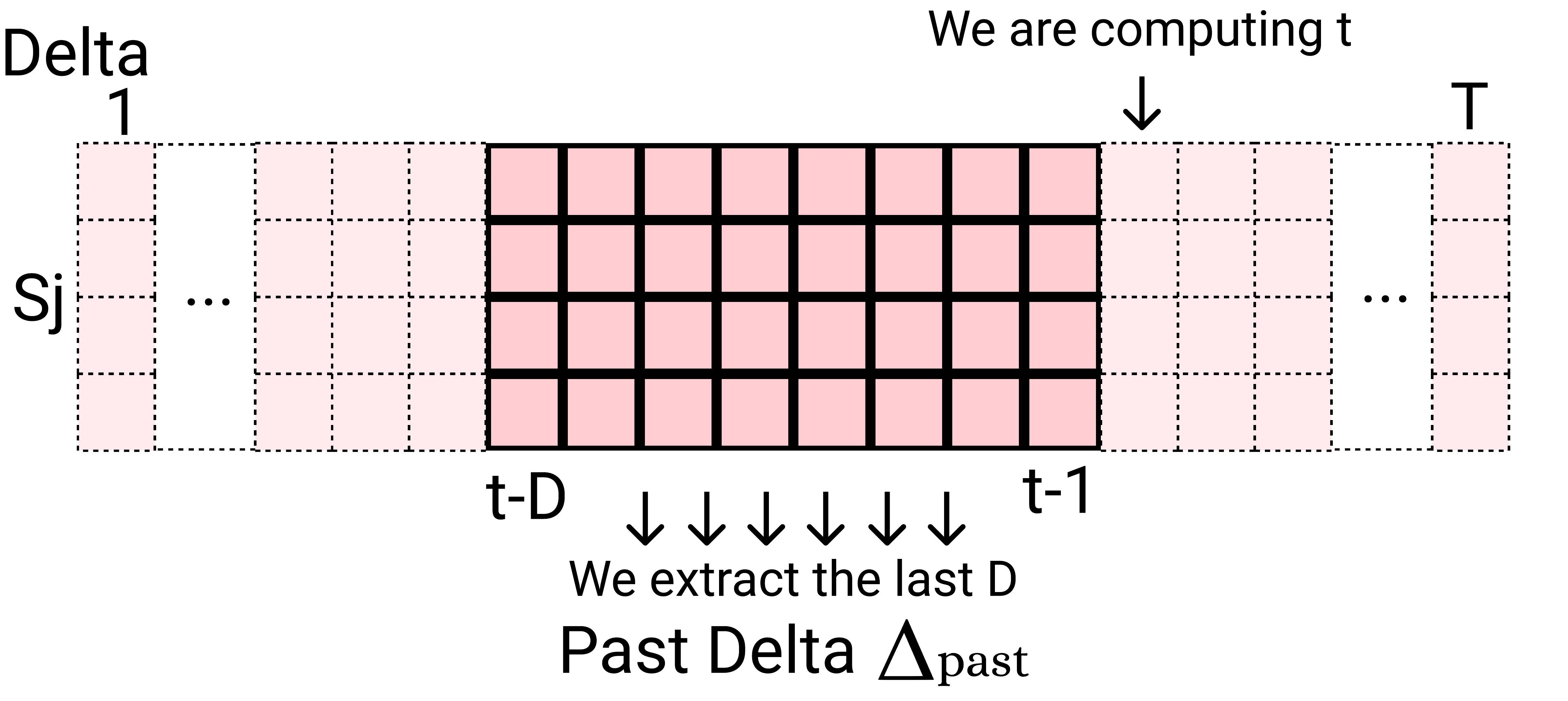}
    \caption{Extracting from $\boldsymbol{\Delta}$ past $[t\!-\!D,t)$ values to obtain $\boldsymbol{\Delta}_{past}$.}
    \label{fig:extracting_pastdeltas.png}
\end{figure}

The past delta values are the simpler of the two time-dependent matrices.
We collect from the stored $\boldsymbol{\Delta}$ matrix a window of size $N \times D$ obtaining $\boldsymbol{\Delta}_{past}^{N\times D}$.

\begin{equation*}\label{eq:delta_matrix}
    \boldsymbol{\Delta}_{past}^{N\times D}
    \;=\;
    \begin{bmatrix}
        \delta_1(t-D)   & \cdots & \delta_1(t-1) \\[2pt]
        \delta_2(t-D)   & \cdots & \delta_2(t-1) \\[2pt]
        \vdots        & \vdots        & \vdots        \\[2pt]
        \delta_N(t-D)   & \cdots & \delta_N(t-1)
    \end{bmatrix}.
\end{equation*}
As shown in Figure~\ref{fig:extracting_pastdeltas.png}, this window starts at $t - D$ (or $1$ if $t < D$) and ends at $t - 1$, reversed so that the first column corresponds to the most recent past step (duration~$d = 1$).
The resulting matrix $\boldsymbol{\Delta}_{past}^{N\times D}$ is aligned on the $x$- and $z$-axes of the \textit{Brick} (lower face).

\subsubsection{\textbf{Computing Emission Probability Matrix}}\label{ssec:emission}

The emission probability matrix is the more complex of the two factors.
Recall from the term (d) of Equation~\ref{eq:induction_delta} that, for a given state~$s_j$ and duration~$d$, we need the product of the emission probabilities over the $d$~most recent observations. In the sequential algorithm this partial product is trivially computed inside the duration loop.
In the tensor formulation, however, the duration loop has been eliminated.
We therefore need to compute the entire $N\times D$ matrix of cumulative emission at each time step.

\begingroup
\setlength{\arraycolsep}{2pt}
\begin{equation*}\label{eq:emission_product}
    \mathbf{E}^{N\times D} =
\begin{bmatrix}
    b_1(o_{t}) & \prod_{k=0}^{1} b_1(o_{t-k}) & \dots & \prod_{k=0}^{d-1} b_1(o_{t-k})\\[1pt]
    b_2(o_{t}) & \prod_{k=0}^{1} b_2(o_{t-k}) & \dots & \prod_{k=0}^{d-1} b_2(o_{t-k})\\[1pt]
    \vdots & \vdots & \ddots & \vdots & \\[1pt] 
    b_N(o_{t}) & \prod_{k=0}^{1} b_N(o_{t-k}) & \dots & \prod_{k=0}^{d-1} b_N(o_{t-k})
\end{bmatrix}.
\end{equation*}
\endgroup

\begin{figure}[h]
    \centering
    \includegraphics[width=0.9\columnwidth]{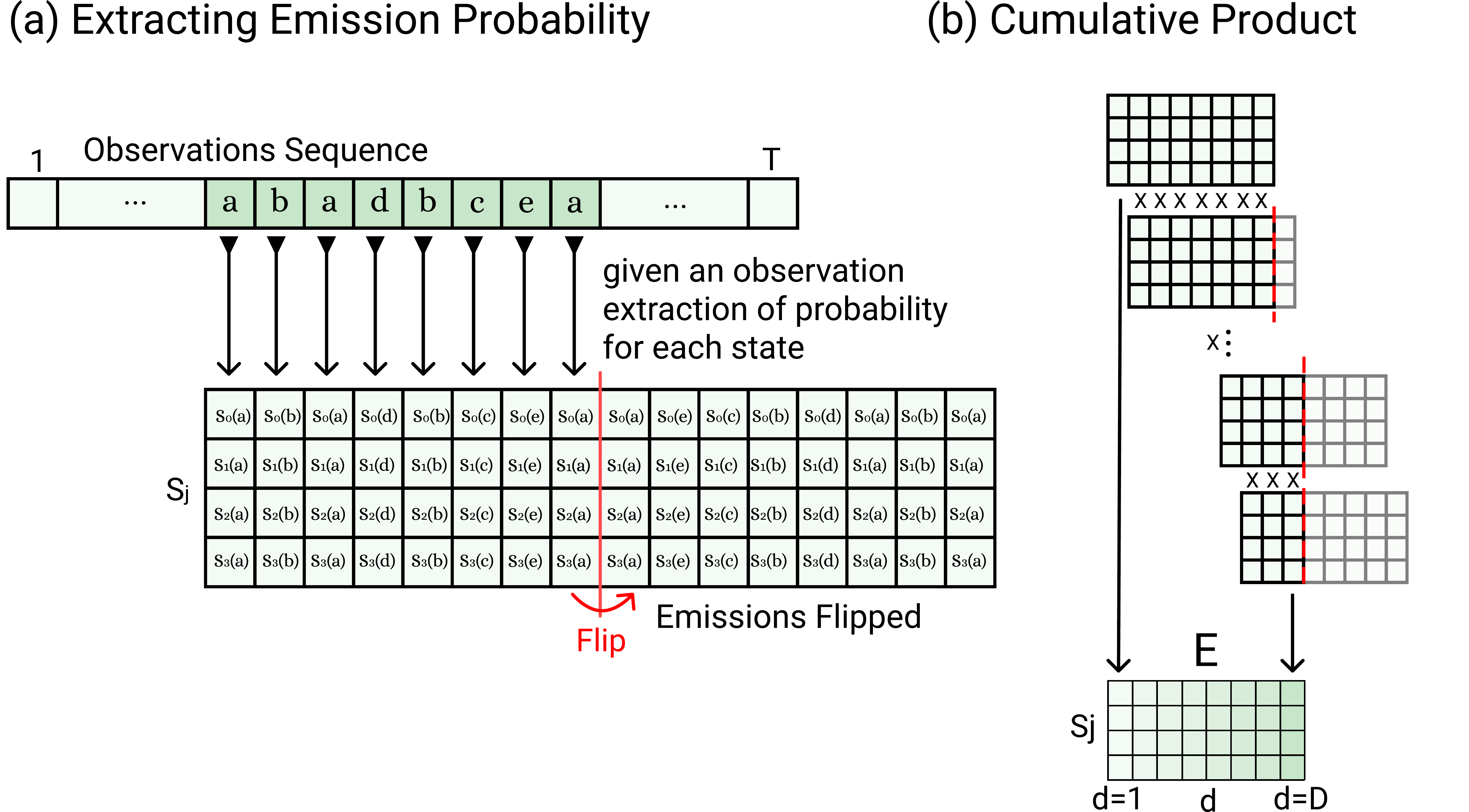}
    \caption{Computing Emission Product to obtain the Emission Probability Matrix.}
    \label{fig:computing_emission}
\end{figure}

As shown in Figure~\ref{fig:computing_emission}, we extract the $D$~most recent observation indices, look up the corresponding emission prob for all states inside $\mathbf{B}^{N \times M}$, reverse the order (so that the first column corresponds to duration~$d=1$). Then, we apply a cumulative product along the duration axis obtaining $\mathbf{E}^{N\times D}$ (Figure~\ref{fig:computing_emission}). 
Our tensor formulation highlights significant redundancies, leading us to propose an optimized caching strategy as detailed in Sec.~\ref{sec:strategies}.

Once the past delta matrix~$\boldsymbol{\Delta}_{past}$ and the emission probability matrix~$\mathbf{E}$ have been computed for a given time step~$t$, we combine them with the precomputed \textit{Brick}~$\mathcal{B}$ through two successive broadcasted products:

\begin{equation}
    \hlmath{brickbg}{ \mathcal{B}_{(t)}^{N\times N\times D}}
    \;=\;
    \mathcal{B}_{first}^{N\times N\times D}
    \;\odot\;
    \underbrace{
        \hlmath{deltabg}{\widetilde{\boldsymbol{\Delta}}_{past~(t)}^{\uparrow\times N\times D}}
    }_{(c)}
    \;\odot\;
    \underbrace{
        \hlmath{embg}{\widetilde{\mathbf{E}}_{(t)}^{N\times \uparrow\times D}}
    }_{(d)}
    \label{eq:step1}
\end{equation}

\noindent

\colorbox{deltabg}{(c)} $\boldsymbol{\Delta}_{past}$ lies on the $x$--$z$ plane, and broadcast the $N\times D$ matrix $N$ times along the $y$-axis, obtaining a tensor $\widetilde{\boldsymbol{\Delta}}_{past}^{N\times N\times D}$. 

\colorbox{embg}{(d)} $\mathbf{E}$ lies on the $y$--$z$ axis, and broadcast the $N\times D$ matrix $N$ times along the $x$-axis, obtaining a tensor $\widetilde{\mathbf{E}}^{N\times N\times D}$. 

The resulting tensor $\mathcal{B}^{N\times N\times D}$ is the fully populated \textit{Brick} for time step~$t$: each entry $\mathcal{B}[j,\,i,\,d]$ encodes the likelihood of transitioning from state~$s_i$ to state~$s_j$ with duration~$d$, given the observations up to time~$t$. Having obtained all combinations of $(s_i,d)$ for each $s_j$, we must now identify the one that yields the maximum likelihood for each $s_j$.

\subsection{Maximum Extraction}\label{sec:max}

\begin{figure}[h]
    \centering
    \includegraphics[width=0.7\columnwidth]{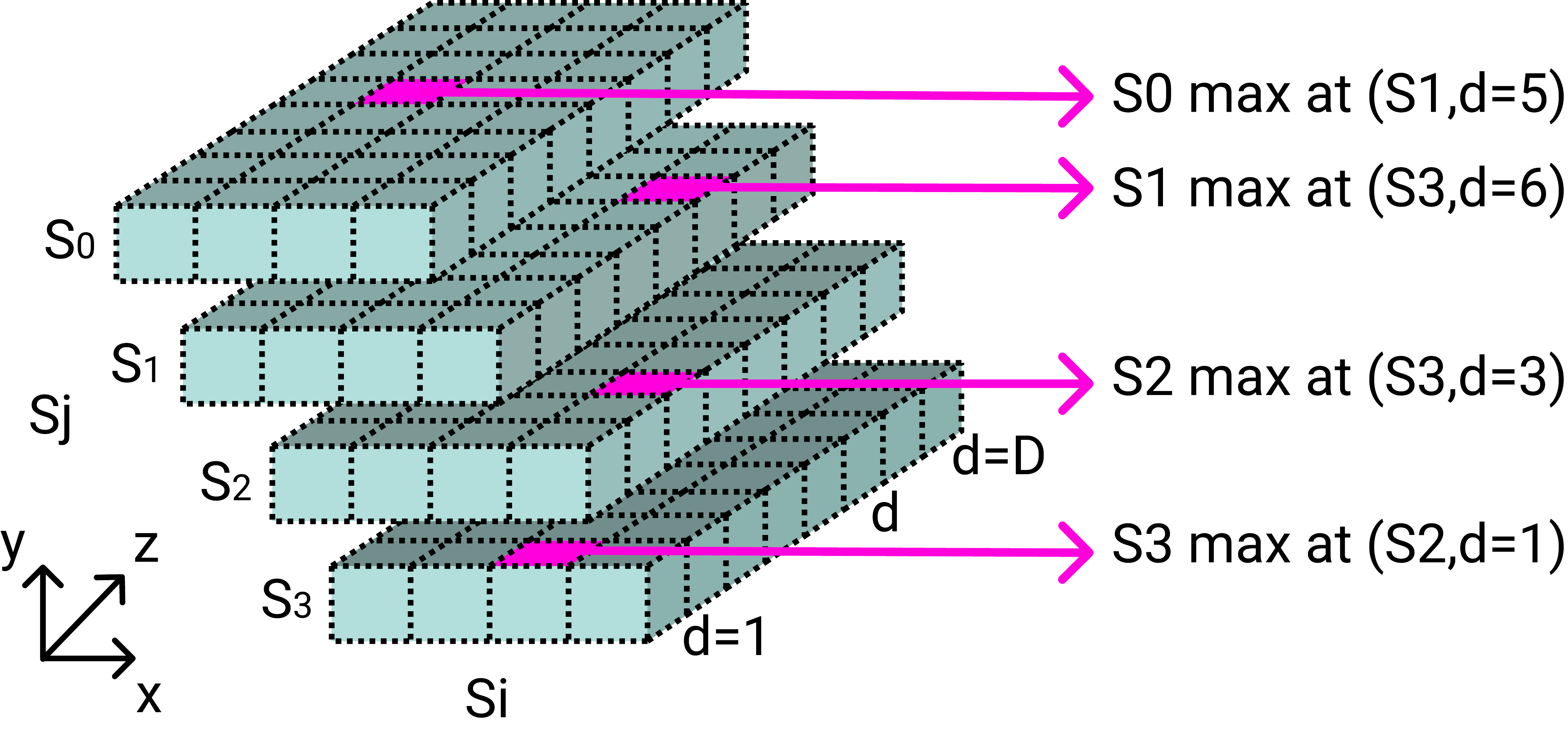}
    \caption{Argmax computation in each $s_j$-slice.}
    \label{fig:slices}
\end{figure}

We now need to extract, for each target state $s_j$, the combination $(s_i^*,\,d^*)$ that maximizes the tensor entry.
Recall that $s_j$ is indexed along the $y$-axis; the corresponding slice is therefore a two-dimensional $N \times D$ matrix.
For each $s_j$-slice we seek the maximum value and its associated coordinates $(s_i,d)$ as shown in Figure~\ref{fig:slices}.
This yields the resulting maximum values vector $\mathbf{\Delta}(t)^N$. This phase corresponds to lines~8--9 of Algorithm~\ref{alg:tensor}.

Seeking the maximum value is a fundamental problem in parallel computing and several strategies can be employed to improve its performance; we discuss our strategy in Sec.~\ref{sec:strategies}.

\subsection{Initialization Phase}\label{sec:init}

As described in Sec.~\ref{sec:seq_init}, the initialization phase covers the first $D$ time steps ($1 \leq t \leq D$), accounting for the possibility that state~$s_j$ has persisted since $t=0$ with no prior transition. In the tensor formulation, the entire initialization is expressed as a single \textbf{broadcasted product} of three matrices (lines~1--2 of Algorithm~\ref{alg:tensor}):
\begin{equation}\label{eq:init_delta}
    \boldsymbol{\Delta}[\forall j,\,1\!:\!D]^{N \times D} = \hlmath{initbg}{\boldsymbol{\pi}^{N \times \uparrow}} \odot  \hlmath{durbg}{\mathbf{P}^{N \times D}} \odot \hlmath{embg}{\mathbf{E}^{N \times D}}
\end{equation}
Here, \colorbox{initbg}{(a)}~$\boldsymbol{\pi}^{N}$ is the initial state probability vector, broadcasted along the duration axis; \colorbox{durbg}{(b)}~$\mathbf{P}^{N \times D}$ is the duration probability matrix; and \colorbox{embg}{(d)}~$\mathbf{E}^{N \times D}$ is the cumulative emission matrix, where each entry $\mathbf{E}[j, d]$ accumulates the emission probabilities of the first $d$ observations under state~$s_j$. Since no entry depends on any other, all $N \times D$ values are computed with no loop-carried dependencies. During these first $D$ steps, both initialization and induction contribute; the final $\delta_t(j)$ is taken as the maximum of the two (lines~8--9 of Algorithm~\ref{alg:tensor}).

\section{Implementations and Optimizations} \label{sec:strategies}
This section details the optimization strategies for the Tensor-Based Viterbi algorithm from Sec.~\ref{sec:tensor-based}. The tensor formulation naturally enables several optimizations: it exposes data reuse patterns for cache-friendly access and structures computation along independent axes, mapping efficiently onto SIMD and multi-threaded execution models.

We developed four versions of the Tensor Viterbi algorithm: \tenspy, \tensoc, \tensmc and \tensgpu. \tenspy is a direct transcription of Algorithm~\ref{alg:tensor} in NumPy, and it was the first implementation developed.
No low-level optimization is attempted; the implementation serves as a readable reference and validation baseline.
It was necessary to analyze the algorithm, identify optimization opportunities, and test them before delving into the low-level optimized implementations.

We highlight that all
operations are implemented in log-space. Since the quantities
involved lie in the interval $[0,\,1]$, repeated multiplications would quickly lead to numerical underflow. Working in log-space is a standard practice adopted by all major Markov model frameworks~\cite{rabiner, berard2025review}. The practical consequence is straightforward: products become sums, cumulative products become cumulative sums, and broadcasted products become broadcasted sums.

\begin{table}[htbp]
\centering
\caption{Summary of Implementations}
\renewcommand{\arraystretch}{1.4}
\label{tab:implementations}
\begin{tabular}{@{}llp{4cm}@{}}
\toprule
\textbf{Naming} & \textbf{Architecture} & \textbf{Language and Tools} \\ \midrule
\tenspy          & CPU Single-Core     & Python, Numpy       \\
\tensoc         & CPU Single-Core     & C++                 \\
\tensmc       & CPU Multi-Core      & C++, OpenMP         \\
\tensgpu         & GPU                 & CUDA, HIP           \\ \bottomrule
\end{tabular}
\end{table}

\subsection{NumPy Version}
 
\tenspy follows the tensor algorithm precisely and is implemented using NumPy, which allows the algorithm to be expressed directly in terms of tensor operations without requiring further low-level optimizations. The memory layout is the one used by NumPy (row-major, depth-first). The broadcasted sums are expressed in two phases: the 2D matrix is broadcast over the absent axis to obtain a 3D tensor, then the two 3D tensors are summed element-wise. This pattern is applied to both the \textit{Brick Computation} and \textit{Brick Update} phases. The \textit{Past Delta Extraction} phase is performed using NumPy slicing operations, and the argmax is computed using NumPy's \texttt{argmax} function independently for each destination state~$s_j$.

Once this first version was coded, analysis of the prototype revealed an optimization opportunity: the emission accumulation, originally recomputed inside the duration loop at every time-step, can be decoupled from the main recurrence and maintained through a rolling cache. After the warm-up phase ($t > D$), the full emission buffer is obtained by a single element-wise addition and a cache shift, reducing the per-time-step cost from $O(DN)$ to amortized $O(N)$ and eliminating the loop-carried dependency on the duration axis. This optimization has been critical in the C++ versions.

\subsection{CPU Implementation}
\label{sec:cpu_impl}
\tensoc and \tensmc are developed in C++
without relying on any tensor library, since neither BLAS~\cite{blackford2002updated} nor
frameworks such as xTensor~\cite{xtensor} provide broadcasted sums natively.
Implementing the operations explicitly also enabled us to fuse
the \textbf{Brick Update} and \textbf{Argmax} phases, avoiding the need to store
the fully populated \textit{Brick} in memory before computing the maximum.

These versions allow us to exploit the optimization opportunities unveiled by the tensor formulation and produce tuned variants of the proposed algorithm.
 
\subsubsection{\textbf{Memory Access Patterns}}

We introduced two complementary flat layouts designed for spatial locality. The $\Delta_{past}$ buffer uses a time-major layout $(t \cdot N + j)$ so that all states at a given time-step are contiguous. The $\Delta$ and $\Psi$ arrays use a state-major layout $(j \cdot T + t)$ as required by the \textit{Backtracking Phase}. The \textit{Brick} tensor uses a $(j, d, i)$ layout where, for a fixed state $j$, the entire $D \times N$ block is contiguous, fitting in L2 across the duration loop if it is small enough. This is a deliberate choice: the hot inner loop sweeps over states $s_i$ within a fixed $(s_j, d)$ slice, achieving contiguous access.

\subsubsection{\textbf{Cached Emissions Computing}}
\label{subsec:cached}
We build a 2D emission buffer $\mathbf{E}$ indexed by $(d \cdot N + j)$, storing the cumulative sum of emission log-probabilities over each observation window. For $t \leq D$, the cumulative sum is computed from scratch. For $t > D$, the buffer is updated incrementally using an \textit{emission cache} $\mathbf{E}_{cache}$: the new observation log-probability at time $t$ is added to a right-shifted copy of $\mathbf{E}_{cache}$, and the result is saved back to $\mathbf{E}_{cache}$ for the next time-step. This reduces the per-state emission update from $O(D)$ to amortized $O(1)$ after the warm-up phase, and makes the emission values for all $(d, j)$ pairs available independently before the \textbf{Argmax} loop begins.

\subsubsection{\textbf{Fused Brick Update and Maximum Extraction}}

The two most intensive phases of the algorithm can be fused in this implementation, performing a fused \textit{Brick Update}$\,+\,$\textit{Argmax} with no self-transition exclusion (the transition matrix encodes this structurally with zero-values). The inner loop is entirely branchless, using annotated ternary conditional assignments. This branchless pattern allows the compiler to generate predicated instructions rather than conditional branches, eliminating branch misprediction penalties. In the three-level loop $(s_j, d, s_i)$, iterations are fully independent across all three axes, making each axis independently parallelizable or vectorizable.

\subsubsection{\textbf{Multi-Core Implementation}}
The \tensmc implementation parallelizes the single-core version using OpenMP. A single \texttt{\#pragma omp parallel} region spawns a persistent thread team, using implicit barriers between $t$~steps to avoid repeated fork/join overhead.

Pre-computation phases exploit full independence across their iteration
spaces. The \textbf{Brick Construction}, for instance, distributes
$N \cdot D \cdot N$ independent element-wise sums across all three axes
via \texttt{collapse(3)}. Similarly, the \textbf{Cached Emission Computing}
exposes $D \cdot N$ independent work units via \texttt{collapse(2)}.
The key parallelization challenge lies in the fused \textbf{Brick Update
and Maximum Extraction}, which in the single-core version offers only $N$
independent tasks. We decompose it into two phases: Phase~A distributes
work over $(s_j, d)$ pairs, where each thread sweeps over all source
states~$s_i$ to find the local maximum, exposing $N \cdot D$ independent
tasks. Phase~B then reduces over~$d$ per destination state~$s_j$. This decomposition enables full thread utilization even when $N$ alone
is smaller than the available core count.

\subsection{GPU Implementation}

The tensor-based Viterbi algorithm is inherently suited for GPU architectures; therefore, building upon our initial CPU version, we developed what is the first GPU-accelerated Viterbi implementation for Hidden Semi-Markov Models to our knowledge. The \tensgpu implementation is developed in CUDA and ported to HIP via \textit{hipify}~\cite{hipify}, maintaining a single codebase for both NVIDIA and AMD architectures.

The primary computational bottleneck lies in constructing ($\mathcal{B}_{first}$) and updating ($\mathcal{B}$) the \textit{Brick} tensor. To maximize hardware utilization, we employ a fine-grained work distribution where each individual thread is responsible for computing a single \textit{Brick} element. Then, after obtaining the final \textit{Brick} version within each time-step iteration, the threads operating on elements of the same $s_j$-slice must cooperate to perform the maximum extraction.

\begin{figure}[h]
    \centering
    \includegraphics[width=1\columnwidth]{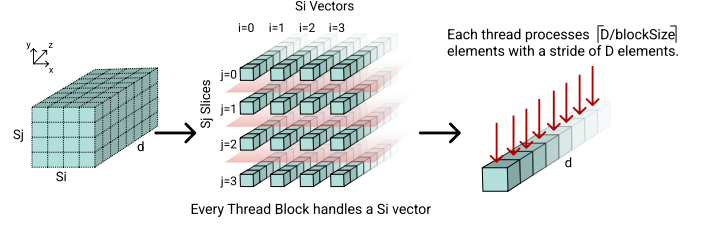}
    \caption{Mapping the \textit{Brick} to GPU Thread Blocks.}
    \label{fig:gpu_mapping}
\end{figure}

To implement this mapping, we decompose the $N \times N \times D$ tensor into $N \times N$ vectors of size $D$ aligned along the $z$-axis, as illustrated in Figure~\ref{fig:gpu_mapping}. This data layout is mapped onto a 2D grid of $N \times N$ thread blocks, where each block manages a single vector. The $D$ elements are evenly divided between the threads in the block. This configuration ensures that each thread block handles an entire temporal slice of the tensor regardless of the duration $D$.

\subsubsection{\textbf{Memory Access Patterns and Coalescing}}
The emission cache (ordered $j \cdot D + d$), the brick $\mathcal{B}_{first}$ (ordered $j \cdot N \cdot D + i \cdot D + d$), and $\boldsymbol{\Delta}$ (indexed $i \cdot T + (t - 1 - d)$) all maintain $d$-contiguity for consecutive and coalesced thread access. Meanwhile, the emission probability $\mathbf{E}$ for the current observation~$o_t$ and 
state~$s_j$ is shared by all threads within a block, allowing 
the GPU to serve it via a single broadcast read.

\subsubsection{\textbf{Cached Emissions Computing}}
Naively computing $\mathbf{E}$ at each time-step requires $O(D)$ work per thread. To eliminate redundant operations, we employ the same cached strategy explored in
Sec.~\ref{subsec:cached}. This reduces complexity to $O(1)$ via a double-buffering scheme. Thread $(j, i, d)$ reads $\mathbf{E}[:,d-1]$ from $\mathbf{E}_{cache}$, adds $b_j(o_t)$, and writes the result to $\mathbf{E}_{final}$. The value is kept in a register for the current iteration, and the two buffers alternate roles across time-steps.

\subsubsection{\textbf{Separated \textit{Brick} Update and Maximum Extraction}}
\label{sec:separated-brick-argmax}

As shown in Sec.~\ref{sec:tensor-based}, the tensor formulation decomposes each time-step into two phases: \textbf{Brick
Update} and \textbf{Maximum Extraction}.  These phases are fused in the CPU
implementation but split across two GPU kernels to better exploit its
architecture.
Note, the time-invariant component $\mathcal{B}_{first}$ is precomputed once
before the induction loop.

\paragraph{First Kernel} each thread $(s_j, s_i, d)$ computes one element of the final \textit{Brick} $\mathcal{B}$ and stores it directly in shared memory, bypassing global memory. An
intra-block parallel reduction then identifies the local maximum score
and its $(s_i, d)$ coordinates in $O(\log D)$ steps: cross-warp steps
use \texttt{\_\_syncthreads()}, while the final $\log_2 (\text{warpsize})$ steps
switch to warp-shuffle instructions (\texttt{\_\_shfl\_down\_sync}),
keeping the running maximum entirely in registers.
\paragraph{Second Kernel} aggregates the per-block local maximums to determine
the global $\arg\max$ over the full $(s_i, d)$ plane for each~$s_j$.

\noindent

Splitting the computation into two kernels avoids grid-wide synchronization, which would otherwise require cooperative-groups designs with additional occupancy constraints.

\section{Experimental Results}\label{sec:exp}
To assess the performance of our tensor-based implementations, we present an in-depth experimental analysis. We first describe the evaluation environment (Sec.~\ref{sec:exp:env}) and validate our implementations (Sec.~\ref{sec:exp:validation}). Then, we evaluate the performance of both CPU (Sec.~\ref{sec:exp:cpu}) and GPU (Sec.~\ref{sec:exp:gpu}) implementations, including an investigation into the impact of varying hardware architectures (Sec.~\ref{sec:exp:arch}). Last, we analyze the energy consumption (Sec.~\ref{sec:exp:energy}) and conduct a stress test using extreme-scale inputs (Sec.~\ref{sec:exp:stress}).

\subsection{Evaluation Environment}\label{sec:exp:env}

\subsubsection{Baseline} 
As a baseline for validation and performance comparison, we selected \texttt{hsmmlearn}~\cite{hsmmlearn}, a C++ library (with a Python API) for HSMMs with explicit duration distributions originating from the R \texttt{hsmm} package~\cite{hsmmr}, making it one of the most established HSMM codebases. We chose it for two reasons. First, it implements the general HSMM formulation with explicit, non-parametric duration modeling and categorical emissions, matching the problem addressed in this work. Second, among the HSMM frameworks analyzed in a recent survey~\cite{berard2025review}, it is one of the few combining a general-purpose formulation with a C++ backend, ensuring our comparison targets optimized compiled code. We refer to this single-core baseline as \baseoc. We also considered \texttt{edhsmm}~\cite{edhsmm}, which is inspired by \texttt{hsmmlearn}, but its Viterbi algorithm is implemented in Cython, potentially limiting low-level compiler optimization compared to a native C++ implementation. 

As \texttt{hsmmlearn} is strictly limited to a single-core implementation, we developed a multi-core variant, \basemc, to ensure a fair comparison. This was achieved by parallelizing the C++ Viterbi decoder with OpenMP, specifically targeting the loop over states, the only loop in the classical four-nested-loops formulation with fully independent iterations. \basemc represents the maximum parallelism extractable from the sequential formulation without significant algorithmic restructuring, serving as our primary baseline for both multi-core and GPU comparisons. It exhibits near-linear scaling with thread count, provided the number of threads does not exceed $N$. Beyond that point, the speedup saturates and degrades slightly due to synchronization overhead, confirming that the traditional formulation fundamentally limits the exploitable parallelism to a single axis. All implementations, including both the baselines and our tensor-based versions, utilize double-precision (FP64) floating-point arithmetic.

\subsubsection{Problem Size} To evaluate our formulation under realistic conditions, we selected problem sizes guided by computational genomics. We tested configurations with a number of states ($N$) ranging from $10$ to $75$, spanning prokaryotic gene finders~\cite{genemark1998} at the lower end, chromatin state annotation~\cite{roadmap2015} in the mid-range ($15$--$25$), and eukaryotic gene finders such as AUGUSTUS~\cite{augustus2003} at the upper end. 

For the sequence length ($T$), we tested from $10^3$ to $10^7$ time steps, covering typical gene-finding invocations~\cite{genscan1997, korf2004} up to $T = 10^6$~\cite{lomsadze2005}. For the maximum duration $(D)$, we adopted values from $100$ to \num{10000}, ranging from the explicit intron duration cutoff of SNAP~\cite{korf2004} to stress-test scenarios capturing the longest gene structure features in the human genome~\cite{sakharkar2004}.

\subsubsection{Architectures}
We evaluated our implementations on a representative set of high-performance CPU and GPU architectures, summarized in Table~\ref{tab:arch}.

\begin{table}[ht]
\centering
\caption{Hardware specifications. SM: Streaming Multiprocessor; CU: Compute Unit; GCD: Graphics Compute Die.}
\label{tab:arch}
\scriptsize
\setlength{\tabcolsep}{4pt}
\begin{tabular}{cllll}
\toprule
& \textbf{Processor} & \textbf{Units} & \textbf{Memory} & \textbf{Compiler} \\
\midrule

\multirow{3}{*}{\rotatebox[origin=c]{90}{\textit{CPU}}}
& AMD EPYC 7A53   & 64\,Cores          & 512 GiB DDR4   & Cray clang v19.0.0 \\
& Grace           & 72\,Cores          & 480 GB LPDDR5x & GCC 14.2 \\
& Xeon 8480+      & 2$\times$56\,Cores & 512 GiB DDR5   & ICX v2024.1.0 \\

\midrule

\multirow{3}{*}{\rotatebox[origin=c]{90}{\shortstack{\textit{NVIDIA}\\\textit{GPU}}}}
& A100 SXM        & 108\,SMs           & 80\,GB HBM2e   & CUDA v12.2 \\
& H100 SXM        & 132\,SMs           & 80\,GB HBM3    & CUDA v11.8 \\
& H200 SXM        & 132\,SMs           & 141\,GB HBM3e  & CUDA v12.4 \\

\midrule

\multirow{2}{*}{\rotatebox[origin=c]{90}{\shortstack{\textit{AMD}\\\textit{GPU}}}}
& MI250X (1 GCD)  & 110\,CUs           & 64\,GB HBM2e   & ROCm v6.3 \\
& MI300X          & 304\,CUs           & 192\,GB HBM3   & ROCm v5.7 \\

\bottomrule
\end{tabular}
\end{table}

 
 
 
 
 
 
 

\subsection{Validation Results}\label{sec:exp:validation}
All implementations produce identical output to \texttt{hsmmlearn} across every tested configuration, achieving $100\%$ decoding accuracy. To ensure exact equivalence, we include the tail adjustment phase without further optimization, matching \texttt{hsmmlearn}'s boundary handling. Notably, \tenspy, our direct NumPy transcription of the tensor formulation (Algorithm~\ref{alg:tensor}), already achieves a $4.5\times$ speedup over \baseoc. Despite comparing interpreted Python against compiled C++, this result demonstrates that the tensor reformulation alone yields substantial gains before any low-level optimization is applied.

\subsection{CPU Speedup Analysis}
\label{sec:exp:cpu}

\begin{figure}[htpb]
    \centering
    
    \begin{subfigure}[b]{0.48\columnwidth}
        \centering
        \includegraphics[width=\textwidth]{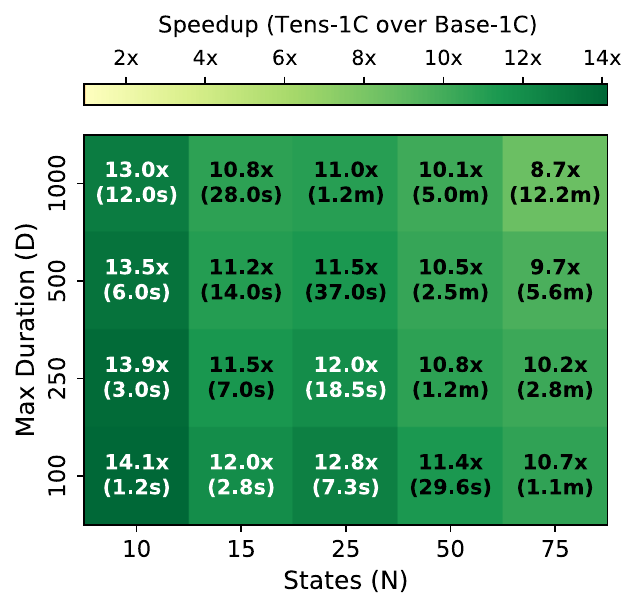}
        \caption{\tensoc speedup over \baseoc and \tensoc runtime (in parentheses; s: seconds; m: minutes). Xeon 8480+, $T=10^5$.}
        \label{fig:heatmap:1c}
    \end{subfigure}
    \hfill 
    \begin{subfigure}[b]{0.48\columnwidth}
        \centering
        \includegraphics[width=\textwidth]{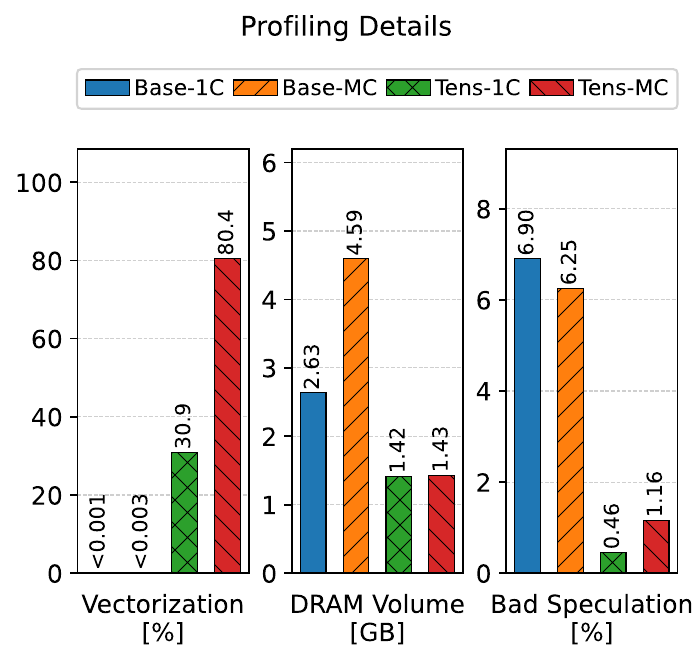}
        \caption{Baselines and Tensors profiling metrics on Xeon 8480+ CPU with ICX compiler.\\ }
        \label{fig:profiling}
    \end{subfigure}
    
    \caption{Speedup and runtime of \tensoc and profiling report.}
    \label{fig:heatmap_profiling}
\end{figure}

We begin by comparing \tensoc against \baseoc on Intel Xeon 8480+. Figure~\ref{fig:heatmap:1c} reports the speedup for $T = 10^5$ across $N \in \{10, 15, 25, 50, 75\}$ and $D \in \{100, 250, 500, 1000\}$.
\tensoc is consistently faster, with speedups ranging from $8.7\times$ ($N\!=\!75$, $D\!=\!1000$) to $14.1\times$ ($N\!=\!10$, $D\!=\!100$). The speedup decreases as either $N$ or $D$ grows, reflecting the point at which $D \times N$ exceeds L2 cache capacity. 

We profiled the execution using LIKWID~\cite{treibig2010likwid}, which reveals that the tensor reformulation reduces retired instructions by $36\times$ (from 610B to 16.9B). Figure~\ref{fig:profiling} highlights the resulting hardware efficiency: the vectorization ratio increases from $<0.001\%$ in \baseoc to $30.9\%$ in \tensoc, while branch misprediction overhead drops from $6.9\%$ to $0.46\%$. The figure also shows a reduced unique DRAM footprint, falling from $2.63$~GB to $1.42$~GB. This, combined with hardware counter evidence of a shift from L2 reuse to streaming L3 access (not shown in the figure), explains both the massive absolute gains and their gradual erosion at larger $N$ and $D$. Nevertheless, even at the largest configuration, the speedup remains significant: \tensoc completes in $12.2$~minutes, whereas \baseoc requires over $1.76$~hours.

\begin{figure}[htpb]
        \centering
        \includegraphics[width=\columnwidth]{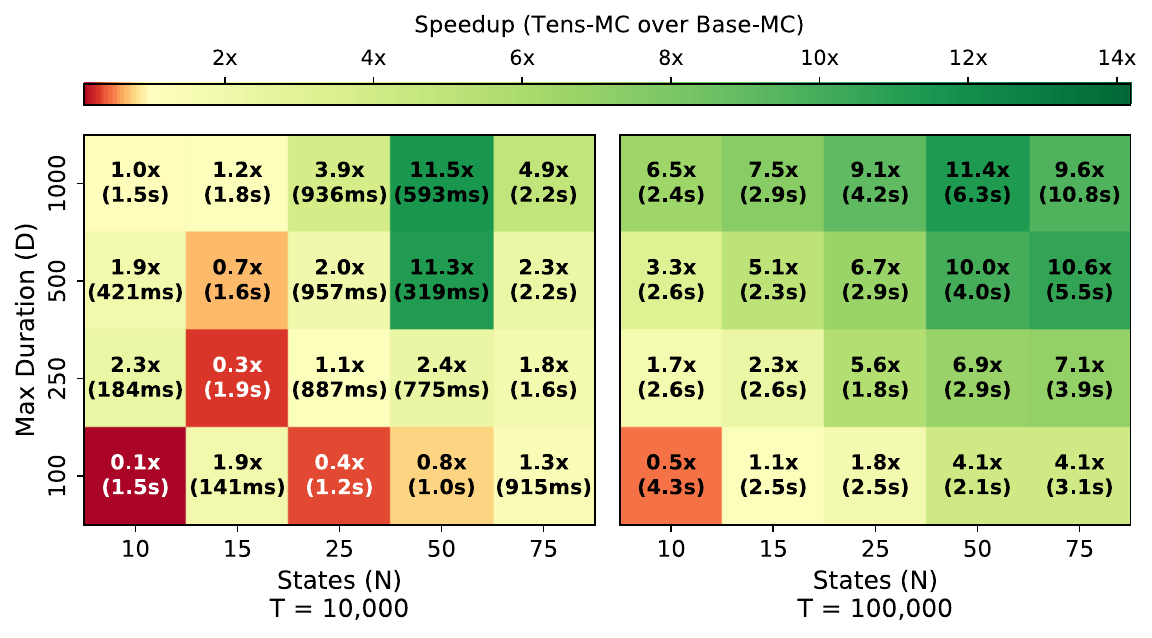}
        \caption{Speedup of \tensmc over \basemc on Xeon 8480+. The cell shows the speedup and the runtime of \tensmc.}
        \label{fig:heatmap_mc}
\end{figure}

We now compare \tensmc against \basemc on Xeon 8480+. Figure~\ref{fig:heatmap_mc} reports the speedup for $T = 10^4$ and $T = 10^5$ across the same value of $D$ (duration) and $N$ (number of states) considered before.
The trend is clear: \tensmc becomes increasingly effective as $N$ and $D$ grow. At $T = 10^5$ the speedup over \basemc reaches $11.4\times$ ($N\!=\!50$, $D\!=\!1000$), while at $T = 10^4$ a peak of $11.5\times$ is observed at the same configuration. This behavior is the direct consequence of the parallelization strategy. \basemc can only distribute work over the $N$ destination states $s_j$, since the sequential formulation carries loop dependencies across durations and source states. 

In contrast, \tensmc reformulates the computation as broadcasted sums with no loop-carried dependencies, exposing $N \cdot D$ independent tasks. This exposes sufficient parallelism to fully saturate the 112 available threads, even when $N$ alone would leave most cores idle. LIKWID profiling confirms this efficiency: \tensmc retires only 9.9B instructions compared to 21.9B for \basemc, while driving L3 bandwidth to near-peak utilization of the memory hierarchy. Furthermore, Figure~\ref{fig:profiling} shows that \tensmc achieves an $80.4\%$ vectorization ratio (versus $0.003\%$ for \basemc), reduces bad speculation from $6.25\%$ to $1.16\%$, and lowers DRAM volume from $4.59$~GB to $1.43$~GB.

\begin{figure*}[htpb]
    \centering
    
    \begin{subfigure}[b]{0.46\textwidth}
        \centering
        \includegraphics[height=4.48cm]{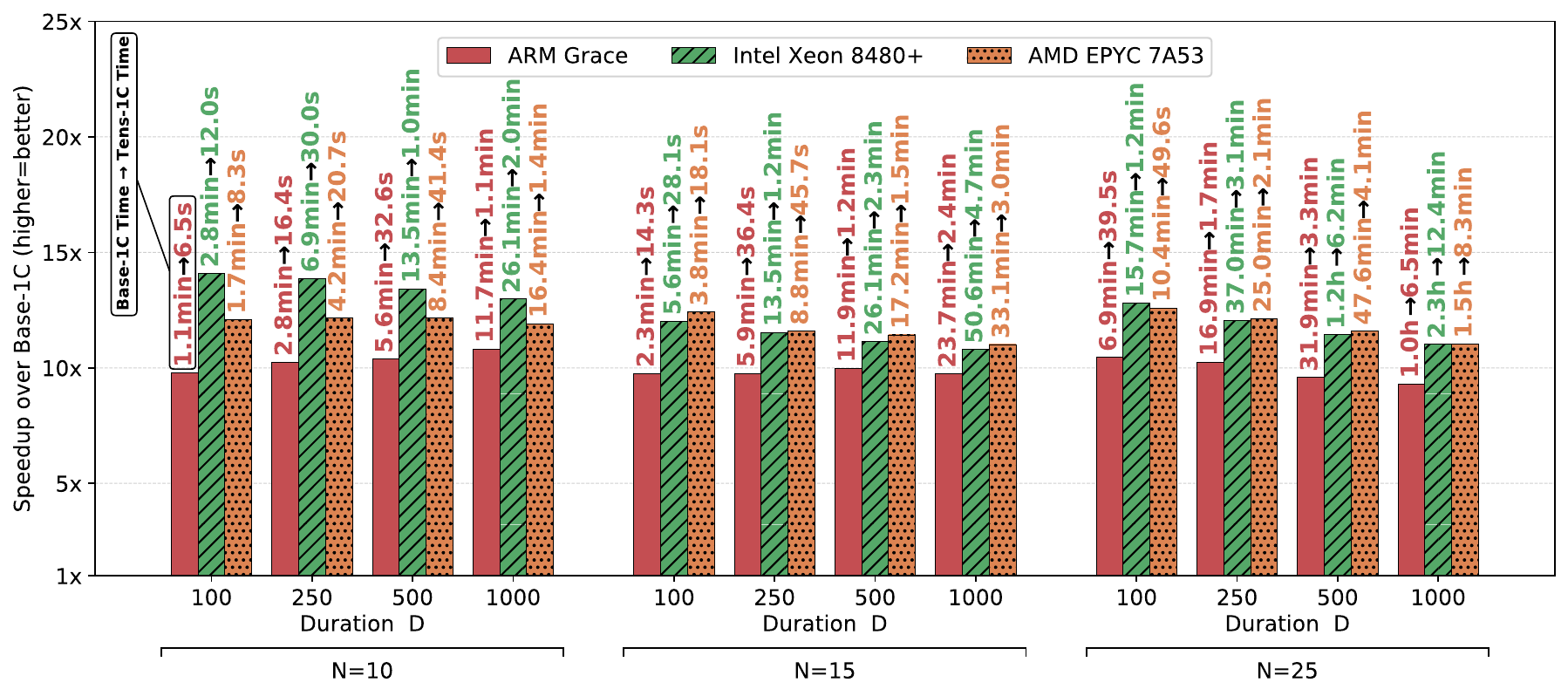}
        \caption{Speedup of \tensoc over \baseoc.}
        \label{fig:arch:oc}
    \end{subfigure}
    \hfill 
    \begin{subfigure}[b]{0.48\textwidth}
        \centering
        \includegraphics[height=4.48cm]{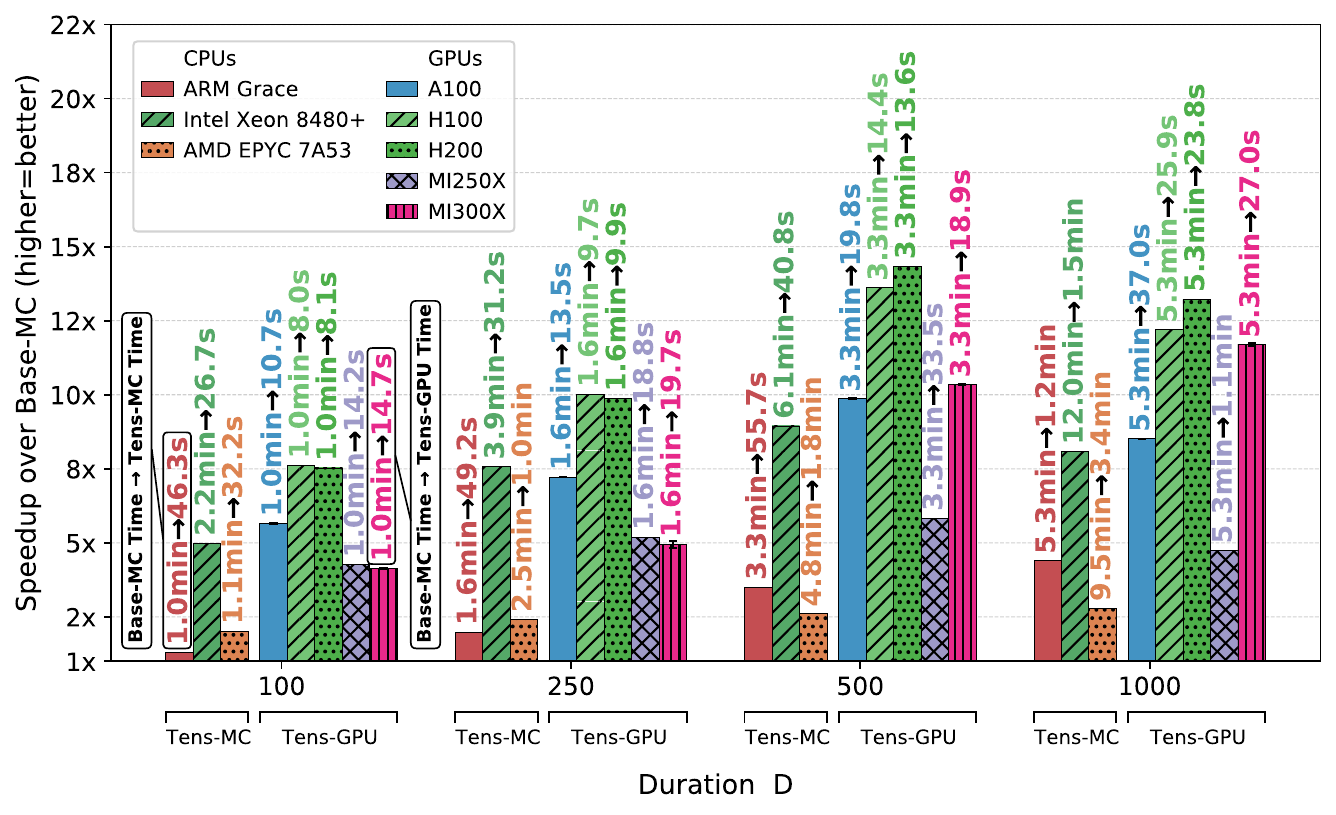}
        \caption{Speedup of \tensmc and \tensgpu over \basemc for N=50.}
        \label{fig:arch:mc} 
    \end{subfigure}
    
    \caption{Speedup of \tensoc, \tensmc, and \tensgpu over different CPUs and GPUs, for T=1,000,000.}
    \label{fig:archs}
\end{figure*}

At small problem sizes, the trend reverses: for $T = 10^4$ with $N=10$ and $D=100$, the \tensmc speedup falls to $0.1\times$. In this regime, pre-computation and barrier overhead dominate because the per-step workload is insufficient to amortize them. This deficit shrinks to $0.5\times$ at $T\!=\!10^5$ and vanishes, becoming a $14\times$ improvement, at $T\!=\!10^6$ (as we will show in Fig.~\ref{fig:arch:oc}), confirming that the overhead is successfully amortized over the longer sequences typical of genomic applications. In Sec.~\ref{sec:exp:arch}, we will evaluate both \tensmc and \tensoc across additional CPU architectures, demonstrating that they consistently outperform their respective baselines. 

Moreover, it is worth remarking that when compiled with GCC, the \tensmc runtime for $D=100$, $N=10$, and $T=10^4$ drops to $0.34$ seconds. This highlights specific inefficiencies in the Intel compiler's OpenMP implementation for small input sizes. Nevertheless, we report all data using the Intel compiler as it consistently outperformed GCC across all other configurations.

\begin{figure}[htpb]
        \centering
         \includegraphics[width=\columnwidth]{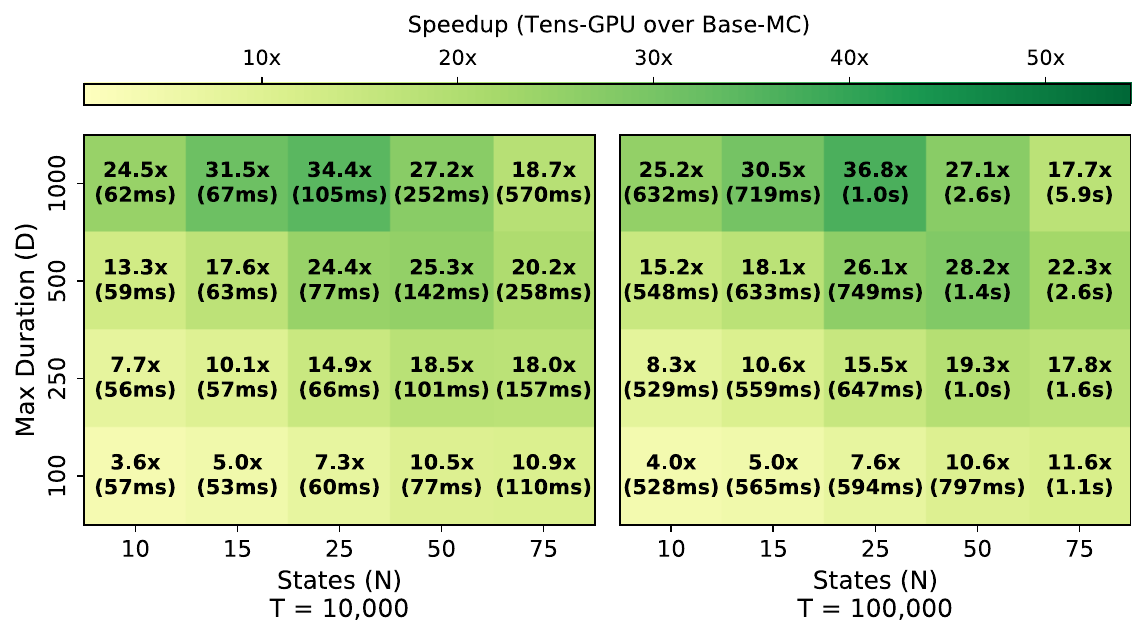}
        \caption{Speedup of \tensgpu on H100 over \basemc on Xeon 8480+. Cells show \tensgpu speedup and runtime.}
        \label{fig:heatmap_gpu}
\end{figure}

\subsection{GPU Speedup Analysis}\label{sec:exp:gpu}
Since no GPU implementation of the HSMM Viterbi exists in the literature we use \basemc, the multi-core variant we developed by integrating OpenMP parallelization into the sequential C++ backend of \texttt{hsmmlearn}, as our reference. We rely on \basemc as a baseline for two reasons: in addition to the total lack of reference GPU implementations, the loop-carried dependencies in the standard sequential formulation across durations and states ($s_i$) fundamentally preclude a direct GPU port of the traditional Viterbi algorithm for HSMMs. Consequently, the algorithmic restructuring proposed in this work is the necessary prerequisite for GPU acceleration. Figure~\ref{fig:heatmap_gpu} reports the resulting speedup of \tensgpu (H100) over \basemc (Xeon 8480+) for $T \in \{ 10^4, 10^5\}$.

\tensgpu achieves speedups across the entire parameter space, peaking at $36.8\times$ ($N\!=\!25$, $D\!=\!1000$, $T\!=\!10^5$). Performance scales with both $N$ and $D$: larger $N$ increases the $N \times N$ thread-block grid, better saturating the H100's 132 SMs, while larger $D$ provides longer reduction vectors per block, improving warp-shuffle efficiency. Even at the smallest configuration ($N\!=\!10$, $D\!=\!100$), the speedup remains $3.6$--$4.0\times$, despite only 100 thread blocks being insufficient to fully occupy all SMs.

For large $N$ ($N\!=\!75$), the speedup plateaus (e.g., $17.7\times$ at $D\!=\!1000$, $T\!=\!10^5$): the per-block shared memory footprint grows with $D$, and the second reduction kernel over $s_i$ states becomes a bottleneck as $N$ increases. 
At $T = 10^6$ the advantage grows further, reaching $54.3\times$ at $N\!=\!15$, $D\!=\!500$: \basemc requires over $5.7$~minutes on 112 CPU cores, while \tensgpu completes the same decoding in $6.4$~seconds. 
Notably, the absolute GPU runtimes remain sub-second for most configurations and never exceed $10$~seconds even at the largest tested ($N\!=\!75$, $D\!=\!1000$, $T\!=\!10^5$), making interactive-scale HSMM decoding on large inputs feasible for the first time. These runtimes open the door to problem sizes that were previously intractable, as we will explore in Sec.~\ref{sec:stress-test}.

\subsection{Architecture Comparison}\label{sec:exp:arch}

Figure~\ref{fig:arch:oc} compares \tensoc against \baseoc at $T = \!10^6$ with $N \in \{10, 15, 25\}$ on the three CPUs described in Table~\ref{tab:arch}. \tensoc delivers consistent speedups of $10$--$11\times$ on Grace, $11$--$14\times$ on Xeon, and $11$--$12\times$ on EPYC across all analyzed configurations, confirming that the gains are portable across architectures. 

In absolute terms, for $N\!=\!25$ and $D\!=\!1000$, \baseoc requires $2.3$~hours on Xeon, which \tensoc reduces to $12.4$~minutes. For $N\!=\!50$ and $N\!=\!75$ (not shown), speedups remain consistent. At $N\!=\!75$ and $D\!=\!500$, \tensoc reduces the runtime from $9.2$~hours to $56$~minutes on Xeon; for $D\!=\!1000$, \baseoc timed out (exceeding $15$~hours), whereas \tensoc completed in $2$~hours. Grace exhibits the lowest runtime, due to its (almost $2\times$) higher memory bandwidth.

Figure~\ref{fig:arch:mc} extends the comparison to \tensmc and \tensgpu at $T\!= \!10^6$ with $N\!=\!50$ across all the CPUs and GPUs introduced in Table~\ref{tab:arch}. For \tensmc, speedups are reported over \basemc on the same CPU. For \tensgpu, speedups are reported over the CPU where \basemc is fastest (Grace).

On the CPU side, at $D\!=\!1000$ \tensmc shows a speedup over \basemc of $2\times$ on EPYC and $8\times$ on Xeon. On the GPU side, the speedup over \basemc grows steadily with $D$, as increasing the duration expands the per-block workload and improves SM occupancy. At $D\!=\!1000$, the speedups over \basemc range from $5\times$ on MI250X to $12\times$ on H200. The lower performance for MI250X can be attributed to its lower memory bandwidth and fewer compute units per GCD. 

Across all GPUs, the absolute runtimes at $D\!=\!500$ remain below $30$~seconds for a million-step sequence. For comparison, the original unmodified \texttt{hsmmlearn} baseline \baseoc requires $3.8$~hours on Grace for this configuration; \tensgpu on a H200 completes the same decoding in $24$~seconds, a reduction of $570\times$.

It is worth mentioning that, to maximize the breadth of our architectural comparison, we opted for a single CUDA/HIP GPU codebase, and OpenMP-only multi-core parallelization. Further specialization is possible on both fronts, and the GPU kernels could exploit architecture-specific features such as distinct memory hierarchies or generation-specific instructions.

\subsection{Energy Consumption}\label{sec:exp:energy}

\begin{figure}[htpb]
    \centering
    
    \begin{subfigure}[b]{0.49\columnwidth}
        \centering
        \includegraphics[width=\textwidth]{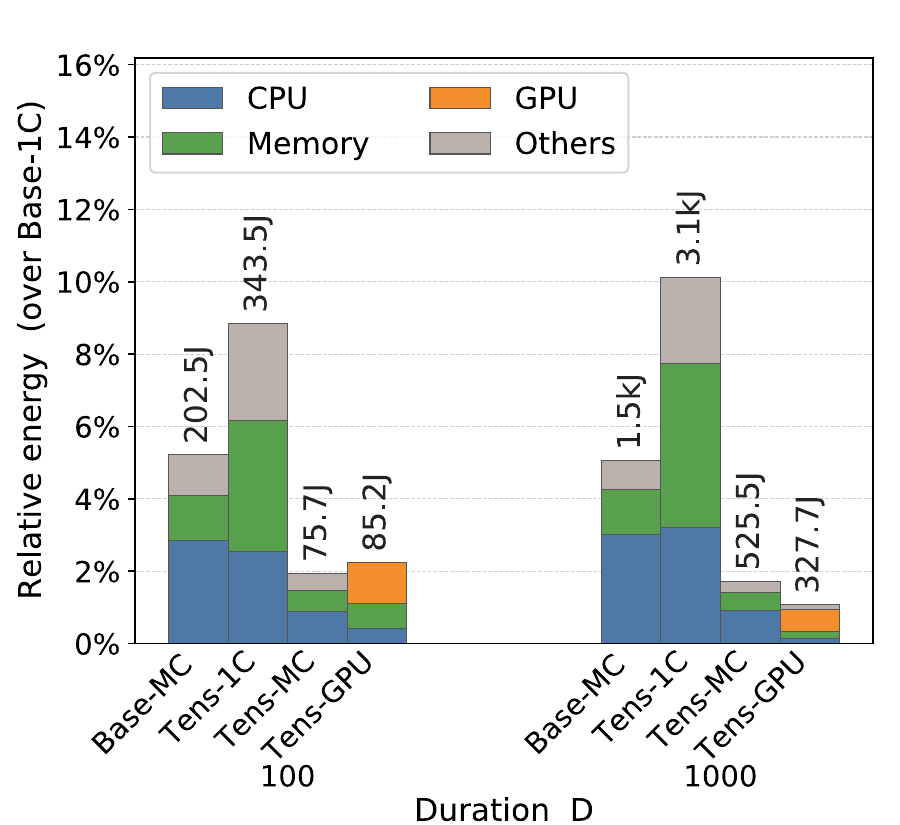}
        \caption{Energy consumption over \baseoc for $N=50$, $T=10^4$ on EPYC 7A53 and MI250X.}
        \label{fig:energy}
    \end{subfigure}
    \hfill 
    \begin{subfigure}[b]{0.49\columnwidth}
        \centering
        \includegraphics[width=\textwidth]{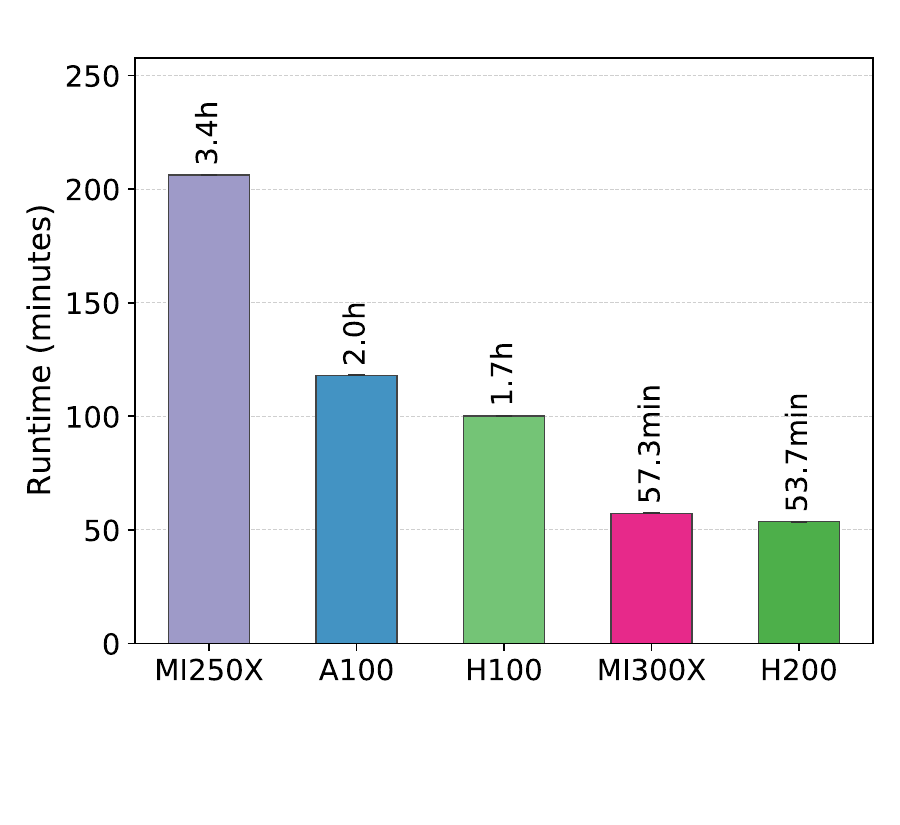}
        \caption{\tensgpu performance on a stress test case with $T=10^7$, $N=100$, $D=10^4$.}
        \label{fig:stress}
    \end{subfigure}
    
    \caption{Energy consumption and stress test.}
    \label{fig:energy_stress}
\end{figure}

Figure~\ref{fig:energy} reports the energy consumption of all implementations at $N\!=\!50$, $T\!=\!10^4$, normalized to \baseoc. For the sake of space, and because we observed a similar trend for the other values of $D$, we only report the data for $D \in \{100, 1000\}$. For the CPU versions, we measure the energy on the EPYC 7A53, whereas for \tensgpu we measure the energy on the MI250X. In both cases, energy is monitored through the Cray Power Management (PM) counters~\cite{craypm}. The dominant factor is execution time: since the instantaneous power draw remains comparable across CPU implementations, energy reductions closely track runtime reductions. 

\tensoc reduces energy consumption by ${\sim}10\times$ over \baseoc across all tested $D$ values, consistent with its single-core speedup. In the multi-core regime, \tensmc consumes $525.5$~J at $D\!=\!1000$, a $3\times$ reduction compared to \basemc ($1.5$~kJ), demonstrating that the tensor reformulation translates its runtime advantage into proportional energy savings. \tensgpu achieves the lowest energy footprint, requiring only $327.7$~J at $D\!=\!1000$ and $85.2$~J at $D\!=\!100$ (just $2\%$ of \baseoc). Although the GPU exhibits higher instantaneous power draw, its shorter execution time more than compensates, rendering it the most energy-efficient platform for $D\!>\!100$. These results confirm that the tensor formulation not only accelerates HSMM Viterbi decoding but also enables a significantly more energy-efficient profile.

\subsection{Stress Test: Beyond Current Workloads}\label{sec:exp:stress}
\label{sec:stress-test}

To demonstrate the practical impact of our formulation, we evaluate \tensgpu on an extreme-scale configuration: $N\!=\!100$ states, $D\!=\!10{,}000$ maximum duration, and $T\!=\!\!10^7$ time steps. This scale is entirely inaccessible to the baseline; \baseoc triggers a memory allocation failure (\texttt{std::bad\_alloc}) for configurations exceeding $T\!=\!10^6$ and $D\!=\!1{,}000$, precluding direct measurement. By extrapolating from runtimes measured at $N\!=\!75$, $D\!=\!100$, $T\!=\!10^6$ (where \baseoc requires  $2$\,h and \basemc $6.7$\,min) using the $O(T \cdot N^2 \cdot D)$ theoretical complexity, we estimate this extreme configuration would require approximately $148$~days for \baseoc and $2.1$~days for \basemc on 112 cores. Such runtimes render not only individual decoding tasks impractical but also make Viterbi training, which requires dozens of such iterations, entirely infeasible on traditional architectures.

Figure~\ref{fig:stress} reports the per-iteration runtime of \tensgpu on five GPUs. The H200 leads  at $53.7$~minutes, followed by the MI300X at $57.3$~minutes, the H100 at $1.7$~hours, the A100 at $2.0$~hours, and the MI250X at $3.4$~hours. The H200 and MI300X's advantage over the H100 is consistent with their higher memory bandwidth and larger number of compute units: at this scale, the $N \times N = 10{,}000$ thread-block grid fully saturates both architectures, and performance becomes bandwidth-bound, favoring the MI300X and H200. The A100 trails the H100 due to its lower bandwidth and fewer SMs. The MI250X, despite its $110$ CUs per GCD, is bottlenecked by its HBM2e bandwidth, the lowest among the five.

These results demonstrate that \tensgpu reduces a previously intractable workload, estimated to take over a month on a single core, to less than an hour on a single GPU, making whole-genome-scale HSMM decoding and iterative Viterbi training practically feasible even for larger sequences.

\section{Related Work}
\label{sec:related_work}
The Viterbi algorithm has been fundamental in high-performance bioinformatics and signal processing for decades, yet existing acceleration efforts are almost exclusively devoted to standard HMMs. Moving from HMMs to HSMMs introduces explicit state-duration handling that substantially increases computational complexity: the Viterbi iteration must compute a maximum over all candidate durations, making the inner loops data-dependent and inherently difficult to parallelize. This combination of computational burden and parallelization difficulty helps explain why performant, hardware-aware HSMM decoders remain absent from the literature.
\subsection{Hidden Markov Models (HMMs)}
HSMMs generalize standard HMMs by introducing explicit state-duration distributions, raising computational complexity from $O(TN^2)$ to $O(TN^2D)$. The Viterbi algorithm has been extensively accelerated for the simpler HMM formulation, including SIMD-vectorized CPU frameworks~\cite{hmmer3, Jiang2018}, CUDA-based GPU implementations~\cite{hoseinyfarahabady2025gpu,yu2014gpu,mohammadidoost2020high,roubtsova2023parallel}, hardware--software co-design and domain-specific approaches~\cite{firtina2024aphmm,hummelgren2024trellis}, and distributed computing~\cite{sassi2021paradist}. However, the additional duration dimension cannot simply be wrapped around existing HMM accelerators: it introduces a cumulative emission product over the $d$ most recent observations, requires accessing a variable-depth window of past delta values, and turns the per-state maximum into a joint maximization over both states and durations. As a consequence, none of these efforts extend to HSMMs, and the HSMM formulation remains entirely unaddressed.
\subsection{Hidden Semi-Markov Models (HSMMs)}
Several statistical frameworks implement major HSMM algorithms in R or Python, with performance-critical routines in C/C++~\cite{berard2025review}. Domain-specific solutions also exist, such as biomvRhsmm~\cite{du2014biomvrhsmm} for genomic segmentation, and Pertsinidou and Limnios~\cite{pertsinidou2015viterbi} that propose Viterbi algorithms based on the backward recurrence Markov chain formulation. All of these implementations, however, are sequential and single-threaded, and none explicitly targets modern high-performance CPUs or GPUs.
The most closely related work is Lu et al.~\cite{lu2025tensor}, who propose a Tensor-based HSMM (T-HSMM) for user activity analysis in Cyber-Physical-Social Systems (CPSSs). Their objective differs fundamentally from ours: their \emph{tensor} refers to embedding multiple correlated entities in a unified higher-dimensional space for activity modeling, rather than targeting computational acceleration, whereas ours reshapes the three inner loops of the Viterbi inductive phase into several 3D tensor operations that expose parallelism for high-performance CPU and GPU execution. Moreover, Lu et al.\ collapse the duration dimension into a single scalar expected value per state, which alters the HSMM semantics and does not solve the exact Viterbi decoding problem. Our formulation instead preserves the full duration dimension and performs exact decoding. A direct head-to-head performance comparison is therefore not meaningful, since the two methods solve different problems.
\subsection{Summary}
To the best of our knowledge, no prior work presents a high-performance implementation of the Viterbi algorithm for HSMMs. Our work fills this gap by proposing a tensor-based reformulation of the HSMM Viterbi algorithm, opening the way to new optimization strategies, accelerator implementations, and application-specific mappings for domains that require Hidden Semi-Markov Model modeling.

\section{Discussion}
We now discuss the main design choices behind our formulation, the trade-offs they involve, and
the technical directions they leave open.

\subsection{Alternative Algorithmic Formulations}
A lower-complexity formulation is in principle available by factorizing the induction, reducing
over source states before combining the duration and emission terms, which lowers the per-step
cost from $O(N^2D)$ to $O(ND+N^2)$. We do not adopt it because the saving in arithmetic is offset
by a loss of hardware efficiency. On CPU, the factorization removes the time-invariant
\emph{Brick} precomputation and with it the contiguous $(j,d,i)$ layout that keeps the working set
resident across the duration sweep. On GPU, it splits a single joint maximization over the
$(s_i,d)$ plane into two reductions that must run one after the other, adding a second grid-wide
synchronization per time step and reducing occupancy, which is exactly the pattern our two-kernel
design avoids. Evaluating this factorization under a different tensor formulation, built around
its own data layout and reduction scheme, would nonetheless be an interesting direction.

\subsection{Mapping onto Specialized Accelerators}
Since emerging AI accelerators and dataflow architectures are designed precisely to execute dense
tensor operations, mapping our formulation onto tensor cores, TPUs, systolic arrays, or FPGA
dataflow designs is a natural direction to consider.
The obstacle is the kind of reduction involved. Since all
quantities are handled in log-space, each step
combines values with an addition and then selects a maximum. These accelerators are instead built
around multiply-accumulate pipelines, so a maximum-based reduction does not map directly onto their native primitives and would need a dedicated mapping strategy. Studying how to support such operations on this class of hardware would
therefore be valuable well beyond our setting, since it would open these units to dynamic
programming algorithms in general. A further consideration is that we use double precision to
match the baseline exactly, whereas peak throughput on these units is available only at lower
precision, so any port must first verify that the dynamic range of the problem allows a narrower
format.

\subsection{Sequence-Level and Distributed Parallelism}
Our implementations decode a single sequence at a time, from start to end. A natural extension is
to split a long sequence into chunks, decode them in parallel, and then reconcile the results at
the chunk boundaries. This would also enable multi-node execution, where each node handles a
portion of the sequence and the boundary values are exchanged through collective operations.
Decoding several independent sequences at once is another promising direction, since it would keep
the device busy on small inputs, where a single sequence leaves many units idle.

\subsection{Model Assumptions and Algorithmic Scope}
Our formulation uses one global maximum duration $D$ for all states, which keeps the tensor dense
and the work per thread uniform. Giving each state its own bound $D_j$ would avoid computing
durations that a state can never take, at the cost of an irregular \emph{Brick}. We also assume
discrete emissions, which makes the emission term a simple table lookup and enables our cached
update; continuous densities such as Gaussians would require a different caching strategy. Finally,
the Forward-Backward and Baum-Welch procedures iterate over the same $(s_j,s_i,d)$ combinations
and only replace the maximum with a sum, so they can reuse the same tensor operations and be
accelerated in the same way.

\section{Conclusions}

We presented a tensor-based formulation of the Viterbi algorithm for Hidden Semi-Markov Models
that restructures the three inner loops of the sequential algorithm into dense tensor operations,
exposing optimization opportunities that are inaccessible to the traditional scalar formulation.
Building on it, we delivered optimized single-core CPU, multi-core CPU, and, for the first time
for HSMMs, GPU implementations, released as the open-source library \textit{tensor-hsmm}.%
\footnote{\url{https://github.com/HLC-Lab/tensor-hsmm/}}

Across three CPU and five GPU architectures, our implementations achieve speedups of up to
$14\times$ on a single core, over $200\times$ with multi-core, and over $570\times$ on GPU with
respect to the sequential \textsc{Base-1c} baseline, while producing output identical to
\texttt{hsmmlearn} in every tested configuration. The gains are structural rather than
platform-specific: a direct NumPy transcription of the formulation already outperforms the compiled
sequential baseline by $4.5\times$, and profiling attributes the compiled speedups to a $36\times$
reduction in retired instructions together with vectorization ratios rising from below $0.001\%$ to
$80.4\%$. Because instantaneous power draw is comparable across implementations, these runtime
reductions translate into proportional energy savings, with the GPU version consuming as little as
$2\%$ of the baseline energy. Most consequentially, a configuration estimated to require over a
month of single-core execution completes in under an hour on a single GPU, bringing
whole-genome-scale HSMM decoding and iterative Viterbi training within practical reach and
establishing a new performance baseline for large-scale HSMM inference.

\section*{Acknowledgements}
We acknowledge ISCRA for awarding this project access to the LEONARDO supercomputer, owned by the EuroHPC Joint Undertaking, hosted by CINECA (Italy). We acknowledge the EuroHPC Joint Undertaking, the LUMI consortium, and BSC for granting access to the LUMI and MareNostrum 5 supercomputers. These resources, hosted by CSC (Finland) and the Barcelona Supercomputing Center (Spain), were provided through the EuroHPC Regular Access program.
The authors used Claude Opus 4.6 and Gemini 3 for
editing the paper; all ideas, content, and conclusions are their own.

\bibliographystyle{IEEEtran}
\bibliography{biblio}

\end{document}